\documentclass[sigconf]{acmart}

\usepackage{float}
\usepackage{microtype}
\usepackage{graphicx} 
\usepackage{makecell}
\usepackage{amsmath}

\usepackage{multirow}
\usepackage{subfigure}

\usepackage{algorithm}
\usepackage{algorithmic}

\AtBeginDocument{%
  \providecommand\BibTeX{{%
    \normalfont B\kern-0.5em{\scshape i\kern-0.25em b}\kern-0.8em\TeX}}}

\copyrightyear{2022}
\acmYear{2022}
\setcopyright{rightsretained}
\acmConference[KDD '22]{Proceedings of the 28th SIGKDD Conference on Knowledge Discovery and Data Mining}{August 14--18, 2022}{Washington D.C.}
\acmBooktitle{Proceedings of the 28th SIGKDD Conference on Knowledge Discovery and Data Mining (KDD '22), August 14--18, 2022, Washington D.C.}
\acmPrice{15.00}

\begin{document}

\title{Game of Privacy: Towards Better Federated Platform Collaboration under Privacy Restriction}



\author{Chuhan Wu$^1$, Fangzhao Wu$^2$, Tao Qi$^1$, Yanlin Wang$^2$, Yuqing Yang$^2$, Yongfeng Huang$^1$, Xing Xie$^2$}
\renewcommand{\authors}{Chuhan Wu, Fangzhao Wu, Tao Qi, Yanlin Wang, Yuqing Yang, Yongfeng Huang, Xing Xie}

\affiliation{%
  \institution{$^1$Department of Electronic Engineering, Tsinghua University, Beijing 100084 \\ $^2$Microsoft Research Asia, Beijing 100080, China}
  \country{}
} 
\email{{wuchuhan15,wufangzhao,taoqi.qt}@gmail.com,yfhuang@tsinghua.edu.cn,{yanlwang,yuqing.yang,xingx}@microsoft.com}
\renewcommand{\shortauthors}{Wu et al.}






\begin{abstract}

Vertical federated learning (VFL) aims to train models from cross-silo data with different feature spaces stored on different platforms.
Existing VFL methods usually assume all data on each platform can be used for model training.
However, due to the intrinsic privacy risks of federated learning, the total amount of involved data may be constrained.
In addition, existing VFL studies usually assume only one platform has task labels and can benefit from the collaboration, making it difficult to attract other platforms to join in the collaborative learning.
In this paper, we study the platform collaboration problem in VFL under privacy constraints.
We propose to incent different platforms through a reciprocal collaboration, where all platforms can exploit multi-platform information in the VFL framework to benefit their own tasks.
With limited privacy budgets, each platform needs to wisely allocate its data quotas for collaboration with other platforms. Thereby, they naturally form a multi-party game.
There are two core problems in this game, i.e., how to appraise other platforms' data value to compute game rewards and how to optimize policies to solve the game.
To evaluate the contributions of other platforms' data, each platform offers a small amount of ``deposit'' data to participate in the VFL.
We propose a performance estimation method to predict the expected model performance when involving different amount combinations of inter-platform data.
To solve the game, we propose a platform negotiation method that simulates the bargaining among platforms and locally optimizes their policies via gradient descent.
Extensive experiments on two real-world datasets show that our approach can effectively facilitate the collaborative exploitation of multi-platform data in VFL under privacy restrictions. 

\end{abstract}

%
%

\keywords{Federated learning, Privacy, Game, Collaboration}

\maketitle

\section{Introduction}

Federated learning (FL)~\cite{mcmahan2017communication} offers the potential to learn intelligent models from decentralized data under privacy protection~\cite{cheng2020federated}.
A typical application of federated learning is leveraging multi-platform data for collaborative model training, where the samples are aligned across platforms while different feature fields of the same samples are kept by different platforms~\cite{hardy2017private}.
This paradigm is usually named vertical federated learning (VFL)~\cite{yang2019federated}.
Existing researches on VFL usually assume that all data on each platform can participate in the VFL coordinated by a target platform~\cite{feng2020multi,wu2020privacy}.
However, in real-world scenarios, the amount of data to be involved in VFL may be limited due to privacy concerns and some strict regulations such as GDPR~\cite{voigt2017eu}.
The privacy budgets\footnote{We define a platform's privacy budget as its maximum accumulated amount of data involved in VFL.} of data owners may also constrain their participation in federated learning~\cite{yang2019federated}.
Moreover, existing VFL methods only benefit the platforms with sample labels of target task~\cite{feng2020multi}.
They cannot provide sufficient incentives for other participants that provide data and computing resources~\cite{han2021data}. 
Thus, there is a huge gap between existing research works and real-world applications of vertical federated learning.

\begin{figure}[!t]
  \centering
    \includegraphics[width=0.66\linewidth]{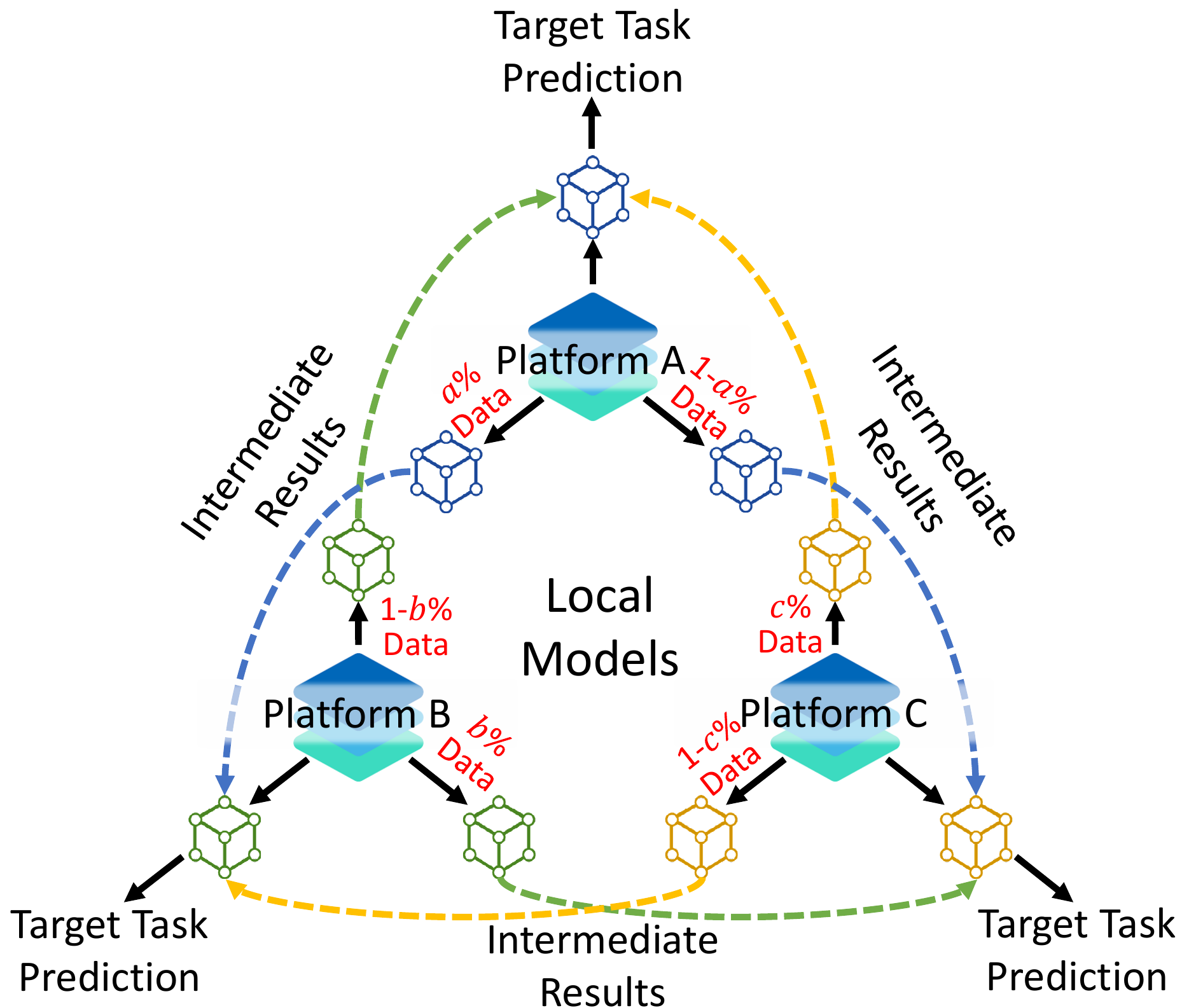}
  \caption{A schematic example of our reciprocal VFL collaboration framework \textit{FedGame}.}
  \label{fig.motivation}
\end{figure}

An intuitive way to incent participant platforms is to allocate profits according to their contributions~\cite{song2019profit}.
Unfortunately, there is no sufficient guarantee that the leader platform is honest and the profit allocation is fair to each platform~\cite{zeng2021comprehensive}.
Instead of currency payment, we propose to incent the VFL participants via reciprocal collaborations~\cite{feng2019joint}, where each platform can benefit its target task from the complementary information encoded in cross-platform data~\cite{roy2014cross}, as shown in Fig.~\ref{fig.motivation}.
Intuitively, a platform can usually better attract another platform's collaboration if it offers more data for federated learning.
However, the privacy budgets of platforms are very limited.
Since collaborating with different platforms may lead to very different returns~\cite{wang2020principled}, platforms need to smartly allocate their data quotas for collaboration with other platforms.
In this way, the reciprocal collaborations among platforms can naturally form a multi-party game, where each platform hopes to obtain a higher model performance gain with possibly less offered data by carefully designing its data quota allocation plan through the game.
The incentives of platforms are obtained by the collaborations and competitions in this game, rather than passively allocated by a leader platform.
The game among platforms can push them to optimize their  collaboration policies, which can improve their ability in  exploiting multi-platform data via VFL.

In this paper, we study the problem of platform collaboration in vertical federated learning under privacy restrictions.
We propose to model the incentive problem in VFL as the aforementioned reciprocal collaboration problem, which can be regarded as a multi-party game with privacy budget restrictions.
We propose a \textit{FedGame}\footnote{Source code is available at https://github.com/wuch15/FedGame.} approach to solve this game.
First, to compute platforms' rewards in the game, we propose a performance estimation method to predict each platform's model performance under different collaboration policies.
More specifically, each platform first offers a small amount of ``deposit'' data to participate in the reciprocal vertical federated learning.\footnote{The intermediate results encoded by the local models and their gradients are exchanged, rather than the raw data.}
Each platform leverages different amounts of inter-platform deposit data to train and evaluate multiple models, and uses multi-variable regression to estimate the real performance gains when involving different amounts of data from other platforms.
Given the performance estimation results, we further propose a platform negotiation method to solve the game by simulating the bargaining behaviors in real-world commercial collaborations that negotiate the ``price'' of provided resources.
Each platform first derives a reward from the expected performance gains and the amounts of data involved in the FL for other platforms, and then performs gradient descent to optimize the reward.
The game among different platforms will converge after multiple rounds of negotiation, and platforms' collaboration policies can be finalized for subsequent collaborative learning.
Extensive experiments on two datasets in different scenarios show that \textit{FedGame} can generate effective platform collaboration policies for VFL to empower the privacy-preserving exploitation of multi-platform data.

The contributions of this paper are listed as follows:
\begin{itemize}
    \item To our best knowledge, this is the first work that studies the competitive collaboration among platforms in vertical federated learning under realistic privacy restrictions.
    \item We propose a performance estimation method to help platforms predict the performance gains under different amounts of inter-platform data involved in federated learning.
    \item We propose a platform negotiation mechanism to solve the multi-party game by simulating the bargaining behaviors in real-world platform collaborations.
    \item We conduct experiments on two datasets for different scenarios to verify the effectiveness of our approach in improving platforms' collaborations in vertical federated learning.
\end{itemize}

\section{Related Work}\label{sec:RelatedWork}

Federated learning is a privacy-aware machine learning paradigm that can leverage highly decentralized private data to learn intelligent models collaboratively~\cite{konevcny2016federated}.
Instead of directly collecting and exchanging private user data for centralized model learning, in federated learning, only the intermediate model updates and variables are exchanged among a server and a number of clients, and thereby user privacy can be protected to a certain extent~\cite{mcmahan2017communication}.
In the standard federated learning, different samples are kept by different clients and their input feature spaces are usually aligned~\cite{li2020federated}. 
This scenario is known as horizontal federated learning~\cite{yang2019federated}, which has been used in various scenarios such as medical data processing~\cite{rieke2020future,kaissis2020secure}, keyboard prediction~\cite{hard2018federated}, and personalized recommendation~\cite{qi2020privacy,yang2020federated,wu2021fedgnn}.

A major variant of the above paradigm is vertical federated learning (VFL)~\cite{zhang2021survey}, which is employed by many methods to exploit multi-platform data for collaborative model training~\cite{yang2019quasi,chen2020vafl,liu2020asymmetrical,fang2021large}.
For example, \citet{hardy2017private} proposed a privacy-preserving logistic regression method that introduces  a third-party server to coordinate the model learning on multi-platform data.
\citet{yang2019parallel} further proposed an improved version of vertical federated logistic regression that is free from the  additional coordinator server.
\citet{wu2020fedctr} proposed a federated ad CTR prediction approach that can exploit user behavior data on multiple platforms in a privacy-preserving way to empower both model training and inference.
\citet{zhang2021secure} proposed an efficient bi-level VFL method that supports asynchronous model updates on all platforms.
In these VFL methods, the samples on different platforms are usually aligned by privacy-preserving entity resolution~\cite{nock2018entity}, while the feature spaces of the same samples are decentralized on different platforms.
Different from horizontal federated learning methods that learn  fully or partially shared models to serve all clients, existing VFL methods mainly benefit the coordinator platform that holds labels in target tasks~\cite{feng2020multi}.
Since other participant platforms need to offer their data and computing resources that lead to additional costs, appropriate incentives mechanisms are necessity in VFL to attract platforms to join the collaborative learning~\cite{zeng2021comprehensive,zhou2021towards}.

\begin{figure*}[!t]
  \centering
    \includegraphics[width=0.85\linewidth]{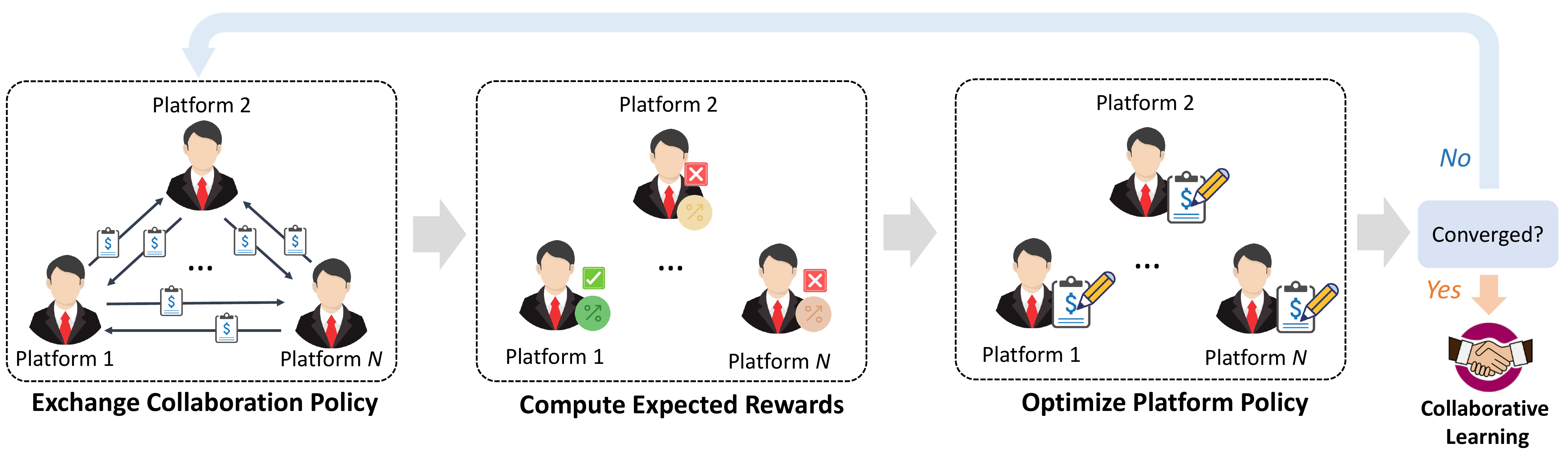}
  \caption{The game framework of our \textit{FedGame} approach.}
  \label{fig.model}
\end{figure*}

Unfortunately, most studies focus on incentives in horizontal federated learning~\cite{song2019profit,lyu2020collaborative,zhan2020learning,yu2020fairness,ng2021multi,yan2021fedcm}, while
the incentive problem in VFL is rarely studied.
Only a few works explore a relevant problem, i.e., the contribution evaluation in VFL~\cite{wang2019interpret,wang2019measure,han2021data,fan2022fair}.
For example, \citet{wang2019measure} proposed to use Shapley value to measure the contribution of each party to VFL.
\citet{han2021data} proposed an information theory-based metric that modifies the Shapley value computation to assess data values of
different platforms from a game-theoretic perspective.
\citet{fan2022fair} proposed a vertical federated Shapley value measurement that can efficiently measure platform contribution under various VFL algorithms.
These methods usually assume that all data on each platform can be involved in collaborative learning.
In fact, due to the restriction of certain data protection regulations, the total amounts of data that can be exploited by federated learning may be strictly limited~\cite{voigt2017eu}.
In addition, the intrinsic privacy risks of VFL~\cite{jin2021cafe,luo2021feature} may also narrow its use given the limited privacy budgets of data owners.
Thus, the collaborations among platforms under privacy restrictions can be quite competitive.
To design profitable collaboration policies that can obtain higher model performance gains, platforms need to accurately estimate the value of incorporating different amounts of inter-platform data, which existing methods cannot achieve.

Different from existing works, we model the incentive problem in VFL as a reciprocal collaboration, where all platforms aim to maximally improve the model performance in their own tasks through VFL collaborations.
Due to the restriction of platforms' privacy budgets, we further form the reciprocal collaboration as a multi-party game.
To measure other platforms' contributions to one platform in the game, we propose a regression-based performance estimation method to measure the expected performance gains when incorporating different amounts of cross-platform data, which is distinct from existing Shapley value-based party contribution measurements that are agnostic to data volume.
In addition, we propose a platform negotiation approach to solve the game and generate beneficial policies to facilitate the exploitation of multi-platform data.
Our approach can facilitate the application of VFL in real-world commercial collaborations among multiple platforms

\section{Methodology}\label{sec:Model}

In this section, we introduce the details of our game-based VFL approach named \textit{FedGame}, which can achieve competitive collaborations among platforms under privacy restrictions to effectively empower their target tasks with limited cross-platform data.
We first introduce the problem formulation and basic assumptions of this work, and then present the details of our method.

\subsection{Problem Formulation and Assumptions}

In this work, we assume that there are $N$ platforms that keep different types of private data, and sample IDs have been aligned across different platforms.
Each platform has a target task, such as news recommendation and Ads CTR prediction, and the sample labels of this task are only kept on this platform.
We define the privacy budget of a platform as the maximum accumulated amount of data it can provide to join in the VFL coordinated by other platforms, which is denoted as $c_i$ for the $i$-th platform.\footnote{Here we assume that the cost is accumulated even if a platform offers the same samples to the learning on different platforms, because sharing hidden information of the same data with different parties will disclose more private information.}
Due to privacy restrictions, the privacy budget of each platform is limited.
To collaborate with other platforms effectively under privacy restrictions, each platform has a policy to allocate different  data quotas for the VFL on different platforms.
We denote the amount of data offered by the $i$-th platform to join the VFL on the $j$-th platform as $c_{i,j}$, which satisfies the constraint $\sum_{j=1}^N c_{i,j} \leq c_i$ ($c_{i,i}=0$).
The policy of the $i$-th platform is defined as the data quota allocations for all other platforms, which is denoted as $\pi_i=(c_{i,1}, c_{i,2}, ..., c_{i,N})$.
The platform reward $r_i$ is defined as the expected model performance gain in the target task on the $i$-th platform, which depends on the amount of inter-platform data provided by other platforms.
The goal of each platform is to maximize its reward under given privacy budget by optimizing its collaboration policy.
In our scenario, the competitive collaborations among platforms naturally form a multi-party game~\cite{WuS99}.
The different platforms are the agents in the game, and their permitted action is adjusting the amounts of data participated in other platforms' VFL.
The platforms can obtain their collaboration policies by solving this game, and they will conduct the final VFL on the allocated data based on platforms' policies.

\subsection{Overall Framework of \textit{FedGame}}

The overall framework of our \textit{FedGame} approach is shown in Fig.~\ref{fig.model}.
It simulates the negotiation behaviors among different platforms in the game.
In each round of negotiation, different platforms first exchange their collaboration policies.
Based on the shared policies, each platform evaluates the value of its own policy by computing the expected reward it can obtain.
Then these platforms locally update their policy to optimize their expected rewards.
If the optimization needs to continue, all platforms return to the collaboration policy exchanging stage and repeat the above steps.
If the policy optimization of all platforms has converged, the platforms reach a consensus about the collaboration plans in VFL, and finally learn their models according to the finalized plans.
We can see that there are two core problems in this negotiation framework:
\begin{itemize}
    \item How to derive the reward of each platform to evaluate its current policy?
    \item How to optimize the platforms' policies to solve the game?
\end{itemize}
To address these problems, we propose an accurate performance estimation method and an effective platform negotiation mechanism, respectively.
We discuss the details in the following sections.

\subsection{Performance Estimation}

\begin{figure}[!t]
  \centering
    \includegraphics[width=0.88\linewidth]{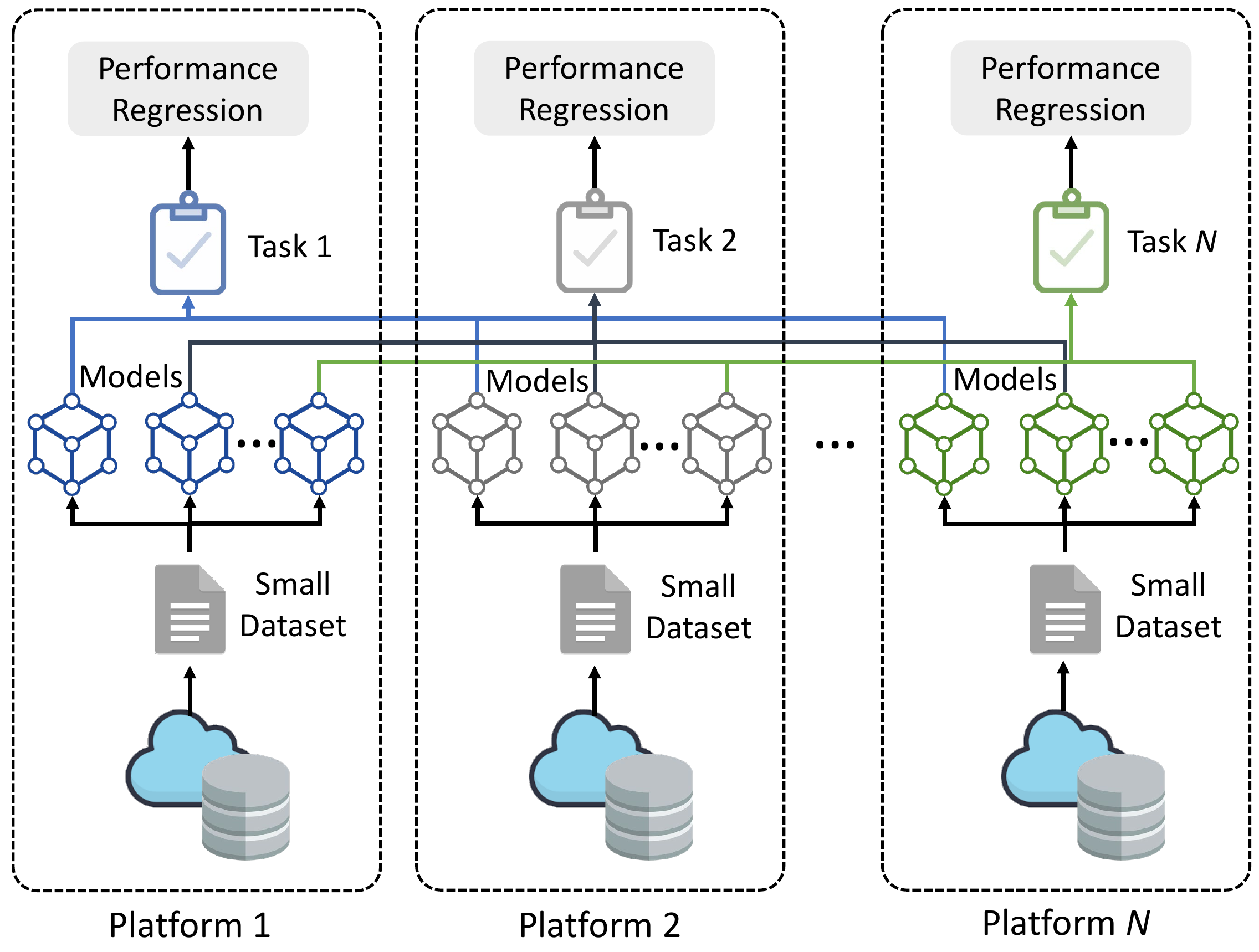}
  \caption{The schematic framework of the performance estimation method in our approach. The repetition of experiments is omitted in this figure.}
  \label{fig.model2}
\end{figure}

In existing methods, the contributions of different platforms are usually evaluated by Shapley value-based measurements~\cite{wang2019interpret,wang2019measure}.
However, Shapley values do not necessarily indicate the real performance contribution~\cite{ma2020predictive,moehle2021portfolio} and they are agnostic to the data amounts.
Thus, they cannot effectively evaluate the real data value with changeable data sizes.
To solve this problem, we propose a simple yet effective performance estimation method.
Its framework is shown in Fig.~\ref{fig.model2}.
The core idea of our method is to learn a performance regression function, which models the relations between the amount of involved data on different platforms and the expected model performance.
Its details are introduced as follows.

In our approach, each platform first offers a small amount (e.g., 5\%) of ``deposit'' data  randomly sampled from the local database to participate in the VFL.\footnote{The deposit data also counts in the privacy budget.}
All platforms use their own local data and the useful information of the deposit data from other platforms to train models for their target tasks, and evaluate the model performance based on their local labeled test data.\footnote{We assume that all local data on each platform is used for learning its own model.}
To accurately model the relation between model performance and the amount of involved data, each platform independently learns and evaluates models  multiple times by randomly sampling different fractions of deposit data from different platforms to participate in the federated learning.
Assume that there are $K$ experiments on the platform $i$, each of which has a data amount combination $(c^k_{1,i}, c^k_{2,i}, ..., c^k_{N,i})$ and a performance observation $r^k_i$.
For this platform, it locally solves a regression problem $\hat{r}_i=f(c_{1,i}, c_{2,i}, ..., c_{N,i})$ based on the $K$ observations, where the independent variables are the amounts of involved data on different platforms, $\hat{r}_i$ is the expected model performance, and $f(\cdot)$ is a regression function. 
Since the model performance is usually better when more data is incorporated, the function $f$ should be monotonic.
Thus, multi-variable linear regression may be a good option due to its simplicity and low requirements of the experiment times $K$ (it only needs to be larger than $N$).\footnote{We do not use more complicated regression methods such as high-order polynomial regression and exponential regression because they have poor accuracy when generalizing to the full data.}
By inputting the $K$ performance observations into the Least Squares Regression method, we can obtain the relations between the observed model performance on each platform and the amounts of involved inter-platform data.
We regress the model performance for all platforms to obtain the following regression functions:
\begin{equation}\label{eq.estimation}
    \begin{aligned}
    \hat{r}_1 & = b_1 + w_{2,1}c_{2,1} + ...  + w_{N,1}c_{N,1},\\
    \hat{r}_2 & = b_2 + w_{1,2}c_{1,2} + ...  + w_{N,2}c_{N,2},\\
    & ...\\
        \hat{r}_N & = b_N + w_{1,N}c_{1,N} + ...  + w_{N-1,N}c_{N-1,N},\\
    \end{aligned}
\end{equation}
where $\hat{r}_i$ is the estimated model performance on the $i$-th platform, $b_i$ is its basic model performance  without inter-platform data, $w_{j,i}$ indicates its relative model performance improvement brought by a unit amount of data on the $j$-th platform.
Using the regression results, we can obtain the estimated model performance gain on each platform given any amount combination of cross-platform data used in model training.

\subsection{Game-based Platform Negotiation}

Based on the performance estimation method introduced above, the platforms can evaluate their rewards given other platforms' collaboration policies.
However, the reward formulation in Eq. (\ref{eq.estimation}) cannot be directly optimized, since a platform cannot control the amount of data offered by other platforms (e.g., platform 1 cannot determine $c_{2,1}$).
Although the amounts of data mutually provided by two platforms are correlated, their latent relations are unknown.
Motivated by the concept of output/input ratio~\cite{chen2002output} in many real-world scenarios, we propose a reward reformulation method to convert platforms' rewards into differentiable forms as follows:
\begin{equation}
    \begin{aligned}
    r_1 & = b_1 + \frac{(w_{2,1}c_{2,1})^\gamma}{c_{1,2}+\epsilon} + ...  + \frac{(w_{N,1}c_{N,1})^\gamma}{c_{1,N}+\epsilon},\\
    r_2 & = b_2 + \frac{(w_{1,2}c_{1,2})^\gamma}{c_{2,1}+\epsilon} + ...  + \frac{(w_{N,2}c_{N,2})^\gamma}{c_{2,N}+\epsilon},\\
    & ...\\
        r_N & = b_N + \frac{(w_{1,N}c_{1,N})^\gamma}{c_{N,1}+\epsilon} + ...  + \frac{(w_{N-1,N}c_{N-1,N})^\gamma}{c_{N,N-1}+\epsilon},\\
    \end{aligned}
\end{equation}
where $\gamma$ is a hyperparameter that indicates the  preference for outputs against inputs, and $\epsilon$ is a small positive constant (e.g., 1e-8) to avoid division by zero.
In these formulas, the reward of a platform is higher if it can use less data to ``barter'' more data on other platforms.
In addition, platforms tend to allocate uniform data quotas (i.e., the same quotas for all other platforms) if $\gamma$ is smaller, while prefer platforms that have greater data contributions to their tasks when a larger value of $\gamma$ is used.

Although $\gamma$ can be empirically tuned, we can have an initial prior estimation of its suitable value to reduce the hyperparameter search effort in practical use.
We can first assume that $c_{i,j}$ is approximately proportional to $c_{j,i}$, which means that contributing more to another platform can obtain more benefits from it.
Since in Eq. (\ref{eq.estimation}), the original reward is also proportional to the data amounts, the value of $\gamma$ should be 2 to achieve the same degree of data amount variables as Eq. (\ref{eq.estimation}).
In fact, due to the constraint of the total privacy budget, the same increase of $c_{i,j}$ may yield a smaller increase in $c_{j,i}$, which means that the correlation between a large $c_{i,j}$ and its associated $c_{j,i}$ may be ultra-linear.
Therefore, the expected value of $\gamma$ is somewhat larger than 2 to achieve the same degree.
In our experiments the optimal $\gamma$ value is about 2.5, which well matches the above analysis.

Finally, we discuss how to solve the game.
In the policy exchange stage in Fig.~\ref{fig.model},  each platform $i$ can only access its concerned collaboration policies (e.g., $c_{j,i}$), but cannot see other irrelevant collaboration policies (e.g., $c_{j,k}$).
Thus, the negotiation is in fact an incomplete information game.
For simplicity, each platform can optimize its own policy based on the information it can access without guessing other platforms' unknown behaviors\footnote{We make this simplification because it can already generate effective solutions.}.
More specifically, each platform uses gradient descent to locally update its policy that can optimize its reward.
The policy $\pi_i$ of the $i$-th platform is updated as follows:
\begin{equation}
    \pi'_i= \pi_i -\eta (\frac{\partial r_i}{c_{i,1}}, \frac{\partial r_i}{c_{i,2}}, ..., \frac{\partial r_i}{c_{i,N}}),
\end{equation}
where $\pi'_i$ is the updated policy, and $\eta$ is a game learning rate.
To ensure the updated policy satisfies the privacy constraint, each of its element $c_{i,j}$ is further normalized as follows:
\begin{equation}
c_{i,j}=
\begin{cases}
  \max(0, c_{i,j}) & \text{ if } \sum_j c_{i,j}\leq c_i, \\
\frac{c_i \max(0, c_{i,j})}{\sum_j c_{i,j}}& \text{ if } \sum_j c_{i,j}> c_i. 
\end{cases}
\end{equation}
It means that all elements are non-negative and their summation does not exceed the budget $c_i$\footnote{We assume that the budget consumption of deposit data has been excluded.}.
After multiple rounds of negotiation in Fig.~\ref{fig.model}, we regard that the game is converged if the policy updates on all platforms satisfy $||\pi'_i-\pi_i||<\mu$, where $\mu$ is a small value indicating the stop criteria.
When the platforms' collaboration policies have been determined, they offer corresponding quotas of data to participate in the VFL launched by other platforms as well as train their own models based on the cross-platform data.

\section{Experiments}\label{sec:Experiments}

To validate the effectiveness of \textit{FedGame} in solving the platform collaboration problem in VFL under privacy restrictions, we design our experiments to address the following research questions:
\begin{itemize}
    \item \textbf{RQ1}: Does the game-based platform collaboration method outperform other baseline strategies?
    \item \textbf{RQ2}: Does the performance estimation method accurately predict model performance under different data quota combinations?
    \item \textbf{RQ3}: What are the collaboration policies obtained by the game solution?
    \item \textbf{RQ4}: How does the output/input preference $\gamma$ influence platform policies and model performance?
    \item \textbf{RQ5}: How does the amount of deposit data influence the collaboration effectiveness?
\end{itemize}
In the following sections, we first introduce the datasets and experimental settings, and then present the experimental results to answer each of the research questions above. 

\subsection{Datasets and Experimental Settings}

We conduct extensive experiments on two datasets.
The first one is \textit{Adult}~\cite{kohavi1996scaling}\footnote{\url{https://archive.ics.uci.edu/ml/datasets/adult}}, which is a widely used tabular dataset.
There are 15 feature fields for each sample.
We divide them into three groups according to their orders in the dataset to simulate the scenario that each platform keeps 5 feature fields of a sample.
On each simulated platform one type of feature is used as the prediction target and the rest are used as inputs (see supplements for detailed feature lists). 
The target tasks on the three platforms are education prediction (16-way classification, platform 1), gender prediction (binary classification, platform 2), and income prediction (binary classification, platform 3) \footnote{The ``education'' and ``education-num'' fields have strong correlations, and we remove the latter field from the input to prevent label leakage.}. 
The second dataset is the Ad CTR prediction dataset used in~\cite{wu2020fedctr}.
We denote this dataset as \textit{CTR}.
Following the settings in~\cite{wu2020fedctr}, we assume that different types of user behaviors in this dataset, including historical clicked Ads, search queries, and browsed webpages, are kept by three different platforms (clicked Ads on platform 1, search queries on platform 2, and browsed webpages on platform 3).
The target task on each platform is to predict whether a candidate Ad will be clicked by a target user based on his/her behaviors.
The statistics of the two datasets are listed in Table~\ref{dataset}.

\begin{table}[h]
\caption{Statistics of the \textit{Adult} and \textit{CTR} datasets.}\label{dataset}

\begin{tabular}{lrlr}
\Xhline{1.0pt}
\multicolumn{4}{c}{\textbf{Adult}}                                \\ \hline
\#training samples & 32,561  & \#test samples            & 16,281 \\ \hline
\multicolumn{4}{c}{\textbf{CTR}}                                  \\ \hline
\#users            & 100,000 & avg. \#queries per user   & 50.69  \\
\#ads              & 8,105   & avg. \#webpages per user  & 210.09 \\
\#ad clicks        & 345,264 & avg. \#words per ad title & 3.73   \\\Xhline{1.0pt}
\end{tabular}
\end{table} 

In our experiments, without loss of generality, we assume the budget of each platform is the amount of the training samples and we use normalized budgets for all platforms, which means that $c_1=c_2=...=c_N=1$ and the data amount variables ($c_{i,j}$) indicate the ratios of budget consumption.
The number of independent experiments for performance  estimation is 5.
We use logistic regression on \textit{Adult} and FedCTR~\cite{wu2020fedctr} on \textit{CTR} as the backbone models.
For the logistic regression algorithm, we use the default settings in Sklearn.
For the FedCTR method we follow its original settings.
In our platform negotiation process, the learning rate $\eta$ is 0.01.
The value of $\gamma$ is 2.5.
The small value $\epsilon$ is 1e-8.
We use accuracy as the metric on \textit{Adult} and use AUC on \textit{CTR}.
We repeat each experiment 5 times and report the average results with standard deviations.

\begin{figure*}[!t]
	\centering
	\subfigure[\textit{Adult}.]{
	\includegraphics[width=0.86\linewidth]{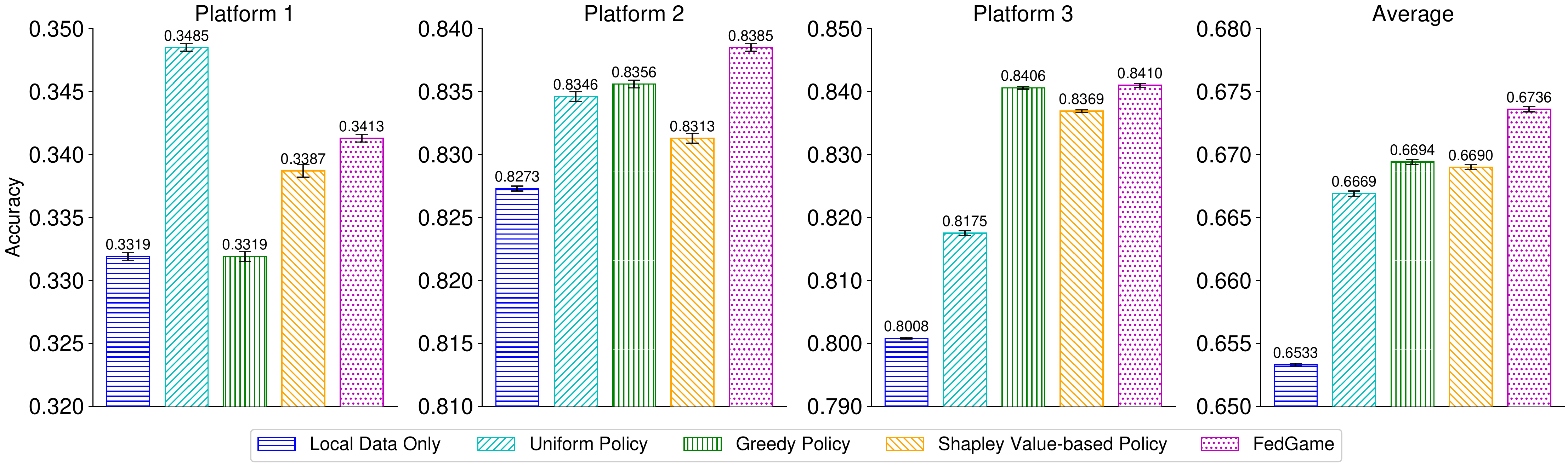}\label{fig.exp1a}
	}
		\subfigure[\textit{CTR}.]{
	\includegraphics[width=0.86\linewidth]{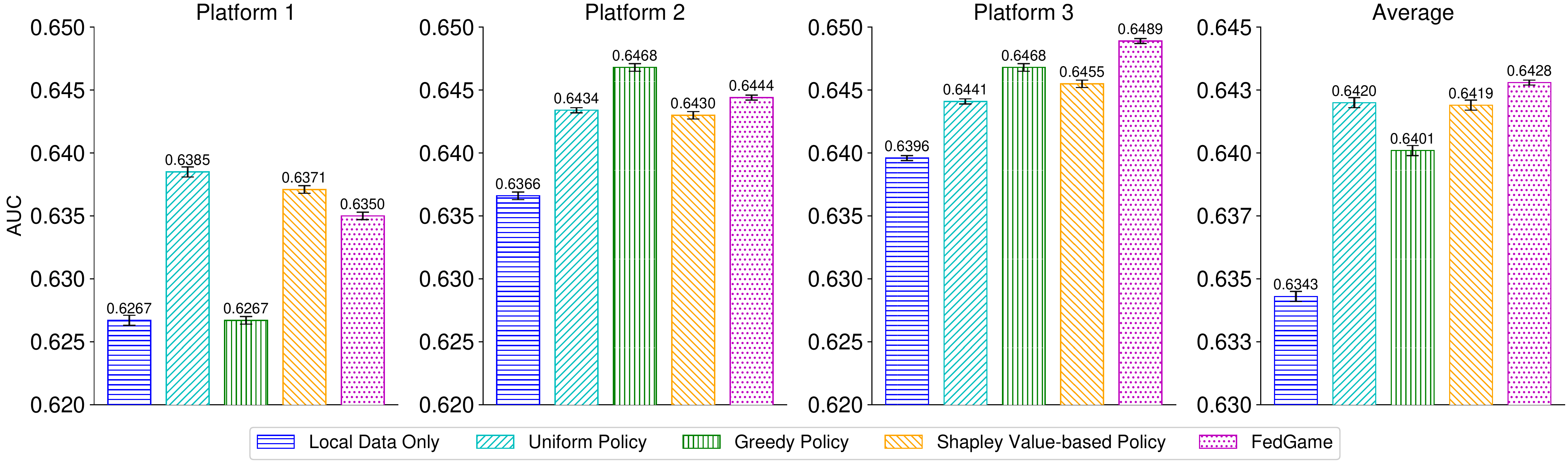}
	}
\caption{Performance evaluation results on \textit{Adult}. Error bars indicate standard deviations.}\label{fig.exp1}
\end{figure*}

\subsection{Performance Evaluation}

To answer \textbf{RQ1}, we verify the effectiveness of \textit{FedGame} in helping platforms' collaborations.
We compare the model performance on each platform as well as the platform average performance using different collaboration strategies, including:
(1) Local data only, where all platforms learn their models solely based on local data;
(2) Uniform policy, where each platform uniformly assigns data quotas for other platforms;
(3) Greedy policy\footnote{We assume that a platform will not offer data to other platforms without returns.}, where each platform chooses to contribute all its data budget to the platform with the highest value to it; 
(4) Shapley value-based policy, which uses deposit data to compute Shapley values and allocate data quotas accordingly.
(5) \textit{FedGame}, our game-based platform negotiation method.
The results are shown in Fig.~\ref{fig.exp1}.
We have several interesting observations.
First, compared with the local data only method, all other methods can improve the average model performance of platforms.
This shows that leveraging multi-platform data can provide rich complementary information to support model training and serving.
In addition, we find that the uniform policy prefers  platforms with low data values (e.g., platform 1 on \textit{CTR} since its model performance based on local data in the same task is the lowest), while is suboptimal for other platforms.
Thus, its average performance is not optimal.
By contrast, we find that the greedy policy makes some platforms excluded from the collaboration (platform 1 does not obtain any data, while platforms 2 and 3 contribute all their data to each other).
Thus, the average performance is also suboptimal.
Besides, the Shapley value-based policy does not perform very well.
This is because Shapley values do not indicate the real performance gains, and allocating data quotas based on Shapley values is not very effective.
Finally, our \textit{FedGame} approach achieves the best average performance (t-test $p<0.01$), and it can even achieve better results than using greedy policy on most platforms.
This shows that using appropriate policies can improve the collaborations between platforms to better exploit multi-platform data.

\begin{figure}[!t]
	\centering
	\subfigure[\textit{Adult}.]{
	\includegraphics[width=0.47\linewidth]{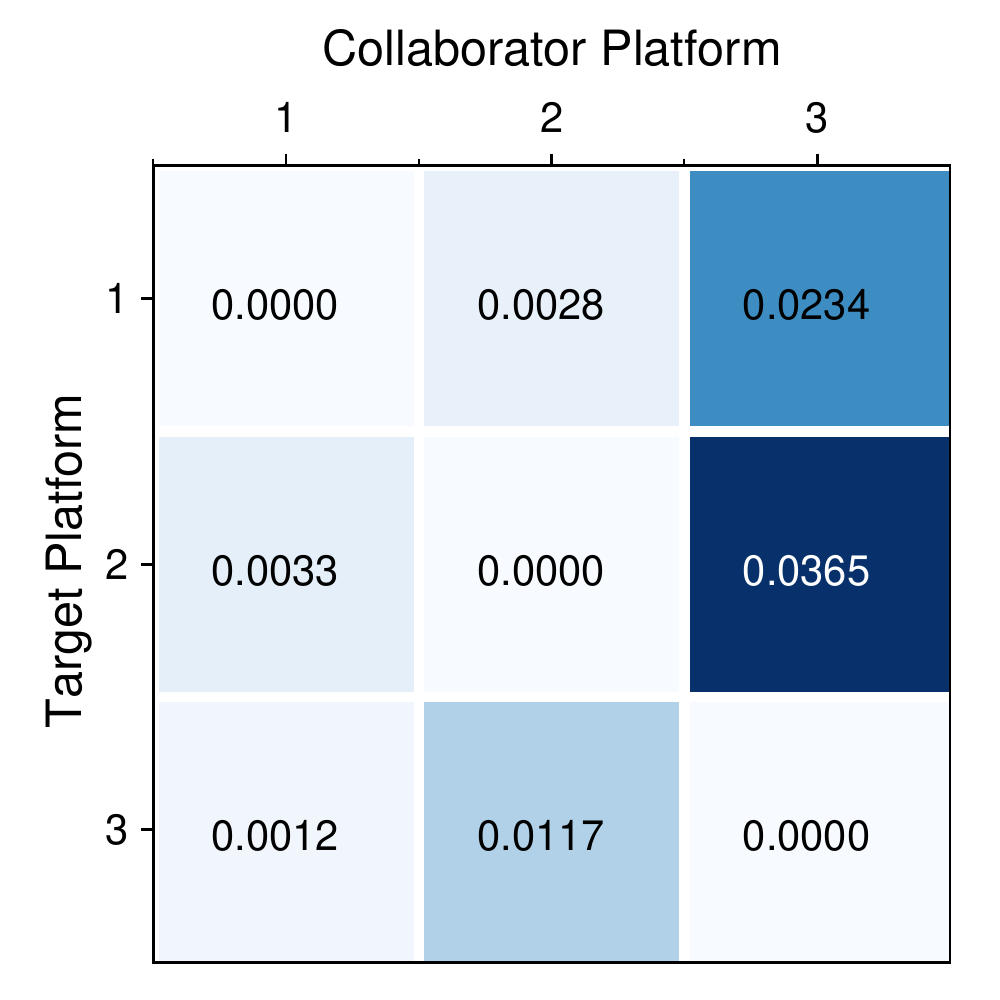}
	}
		\subfigure[\textit{CTR}.]{
	\includegraphics[width=0.47\linewidth]{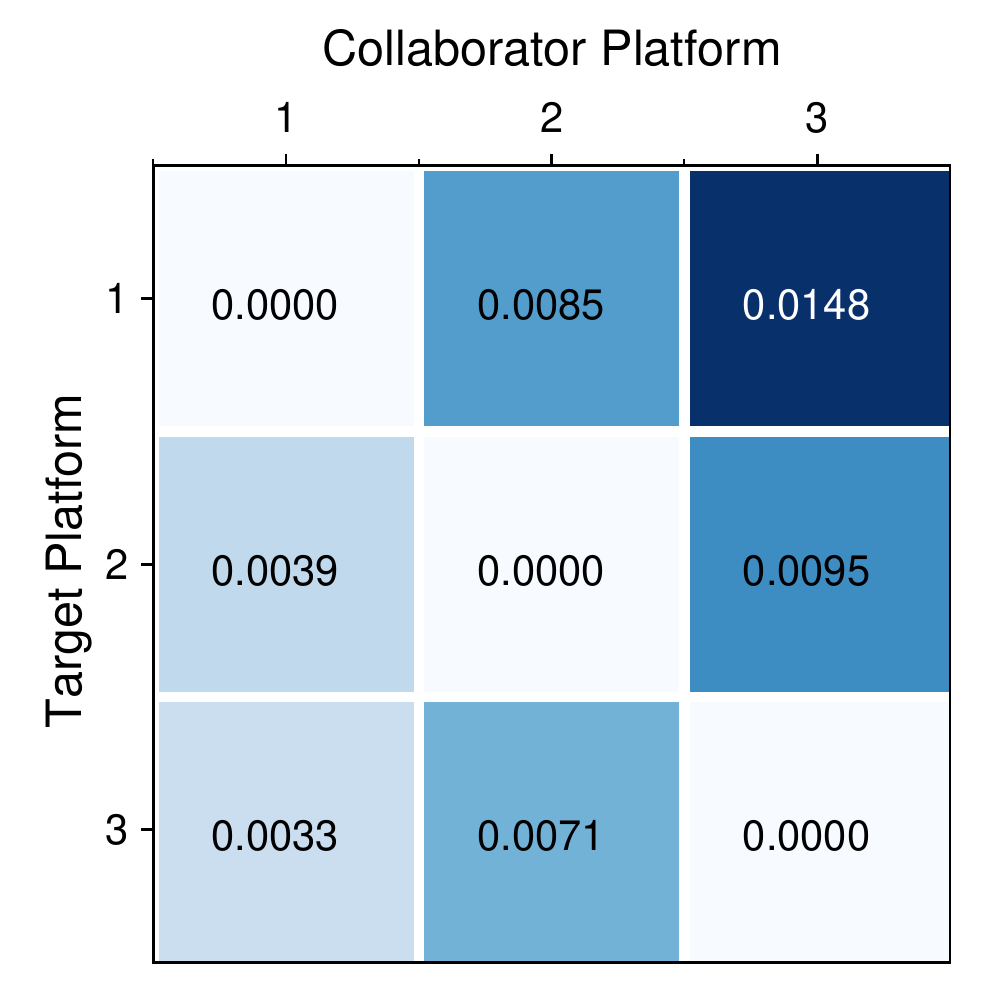}
	}
\caption{Performance regression coefficients on the two datasets. The value at the $i$-th row and $j$-th column is $w_{j,i}$.}\label{fig.exp2}
\end{figure}

\begin{figure*}[!t]
	\centering
	\subfigure[\textit{Adult}.]{
	\includegraphics[width=0.86\linewidth]{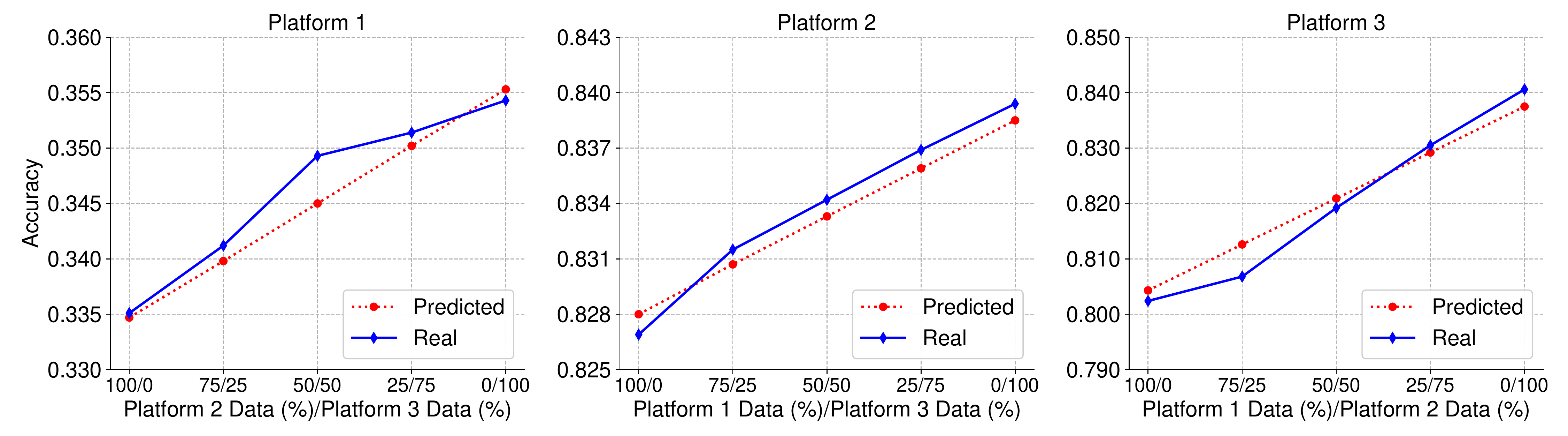}
	}
		\subfigure[\textit{CTR}.]{
	\includegraphics[width=0.86\linewidth]{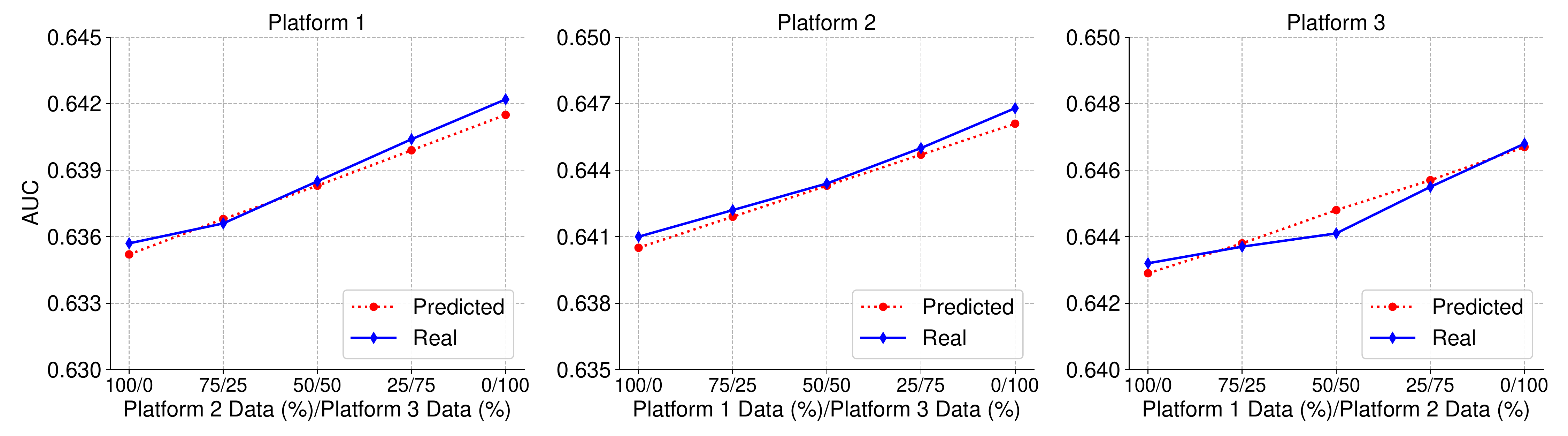}
	}
\caption{Real and predicted model performance.}\label{fig.exp3}
\end{figure*}

\begin{figure}[!t]
	\centering
	\subfigure[\textit{Adult}.]{
	\includegraphics[width=0.47\linewidth]{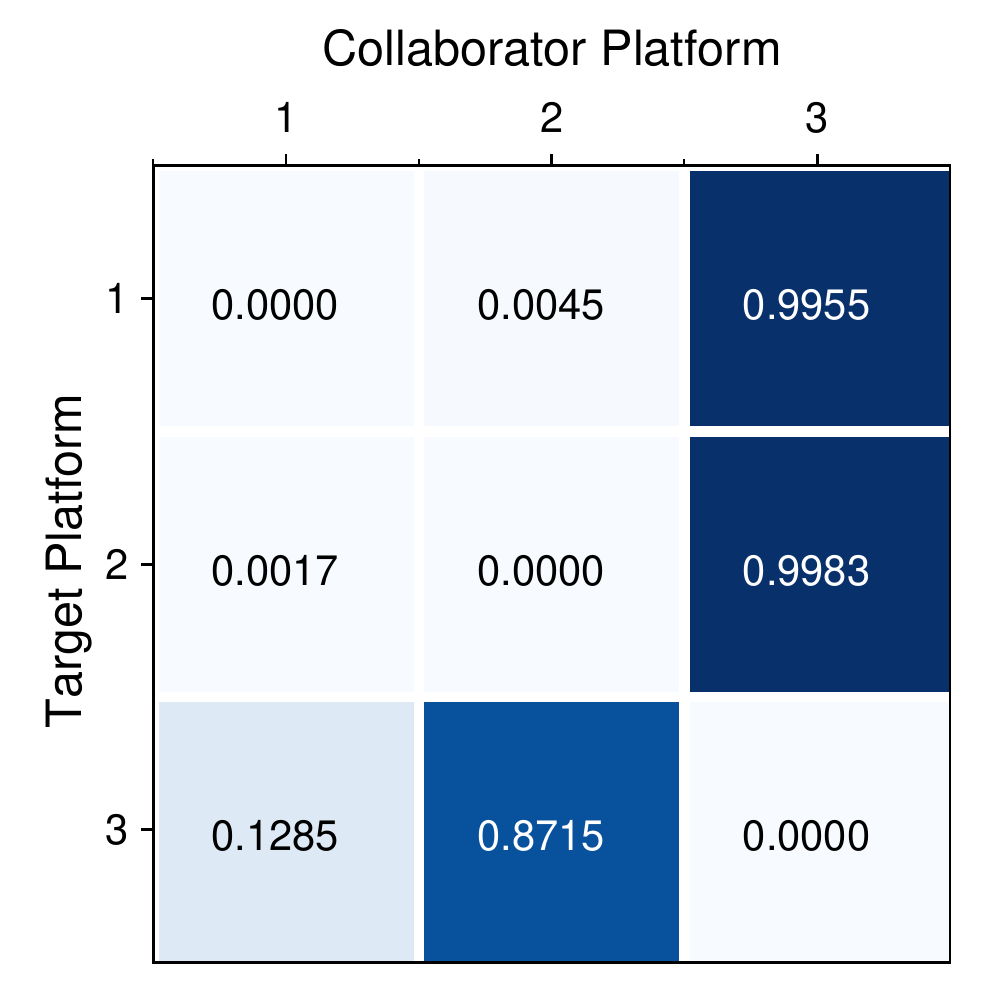}
	}
		\subfigure[\textit{CTR}.]{
	\includegraphics[width=0.47\linewidth]{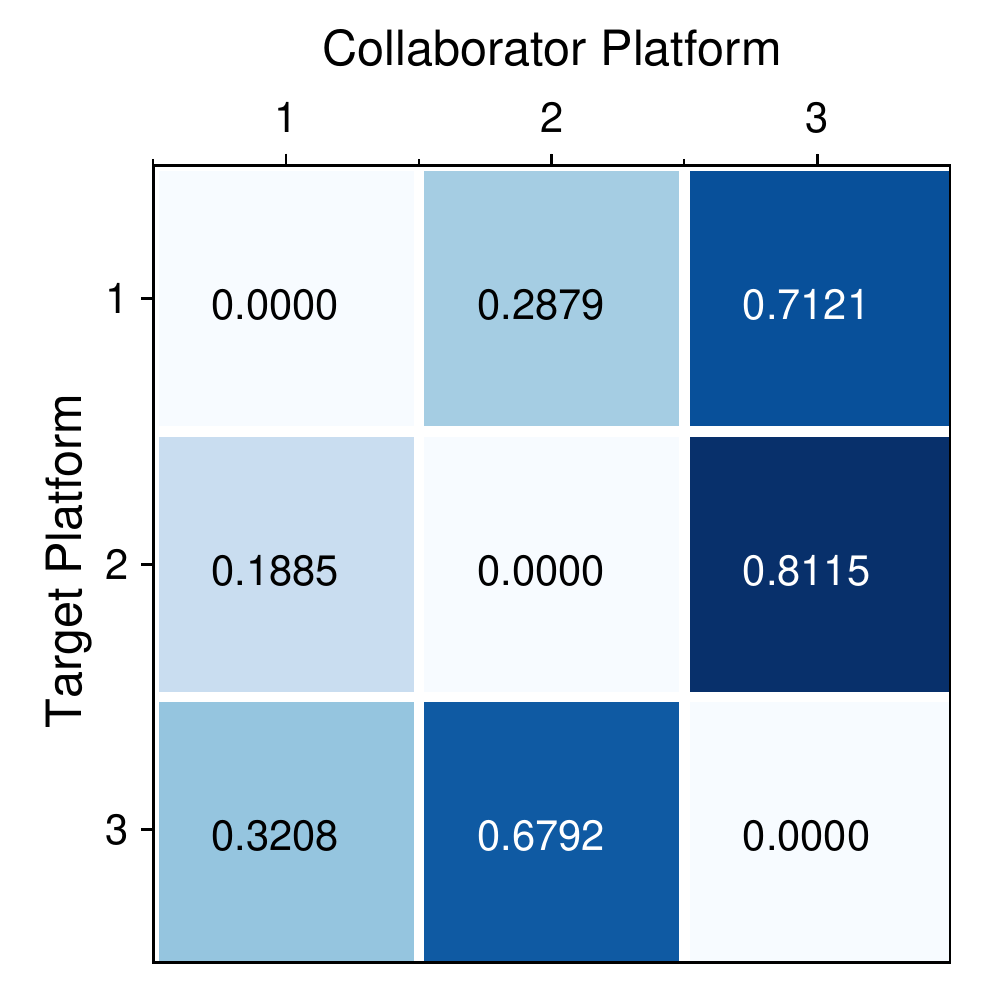}
	}
\caption{The collaboration policies of different platforms. The value at the $i$-th row and $j$-th column is $c_{i,j}$.}\label{fig.exp4}
\end{figure}

\subsection{Results of Performance Estimation}

To answer \textbf{RQ2}, we first show the regression coefficients computed by different platforms in the two datasets in Fig.~\ref{fig.exp2}.
We find although data on different platforms have diverse values for different target tasks, incorporating the data from each platform in both datasets has a positive impact.
Thus, all platforms in both scenarios are qualified to participate in the game.
Since the performance regression is conducted based on the small amount of deposit data, we need to verify its generality to larger amounts of data.
Thus, we compare the predicted and real model performance under different amount combinations of inter-platform data, as shown in Fig.~\ref{fig.exp3}.
We can see that the predicted performance well matches the real performance on all platforms under most amount combinations of involved data.
Thus, if the finally used data amount does not have an extremely huge gap (e.g., 1000 times) with the amount of deposit data, the model performance can be accurately estimated, and the computed rewards can be used as  direct indications of potential model performance gains.

\begin{figure*}[!t]
	\centering
	\subfigure[\textit{Adult}.]{
	\includegraphics[width=0.88\linewidth]{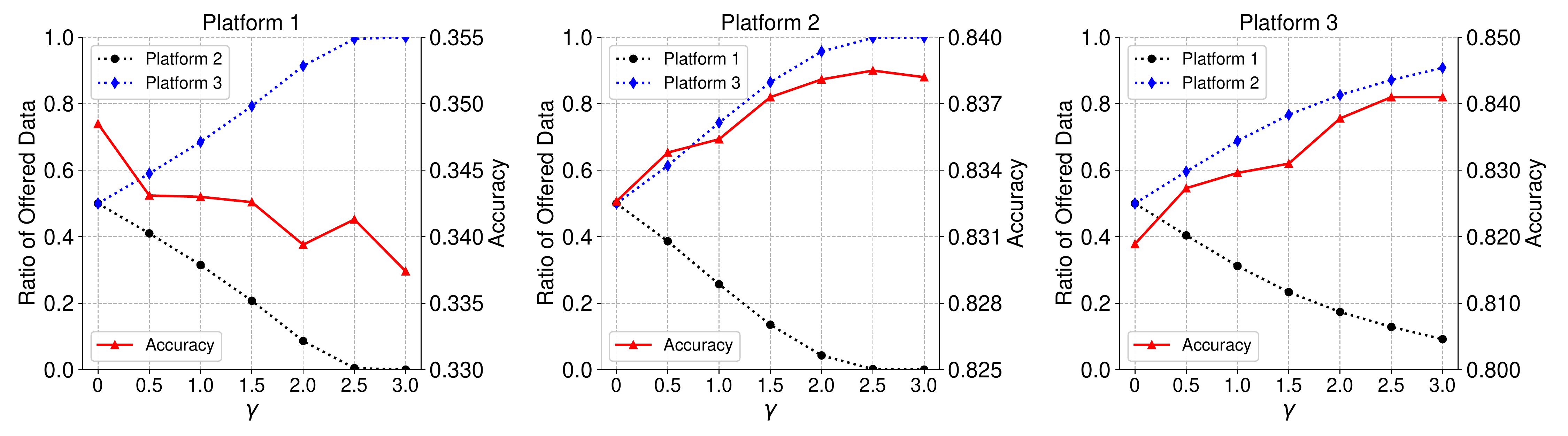}
	}
	\subfigure[\textit{CTR}.]{
		\includegraphics[width=0.88\linewidth]{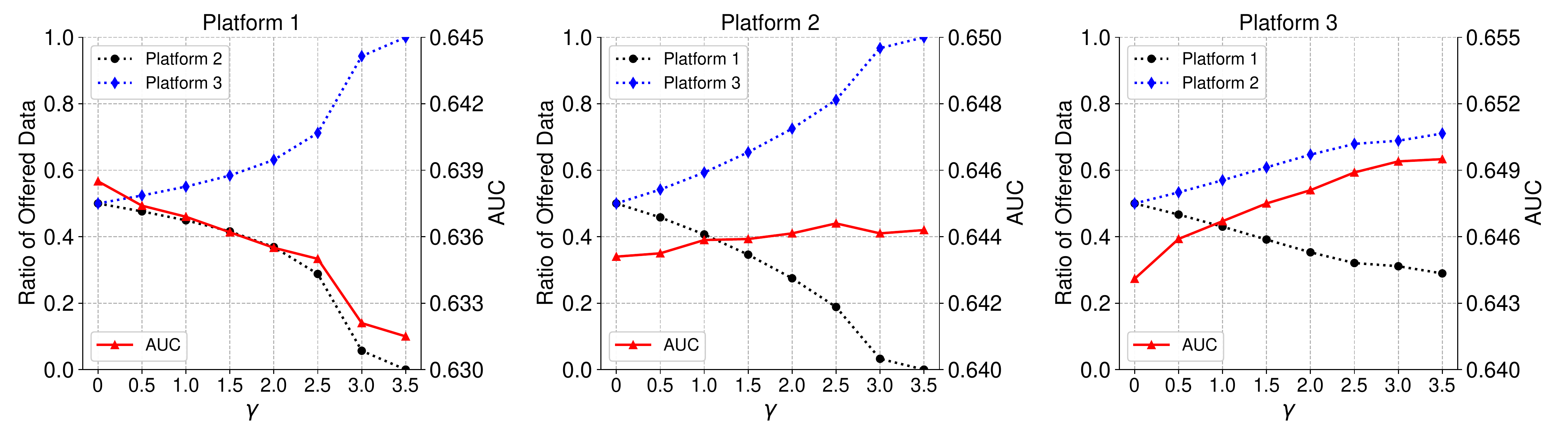}
		}
\caption{The influence of $\gamma$ on platforms' policies and final model performance. The dashed lines  represent the ratios of data offered to other platforms. The best average performance on both datasets is achieved when $\gamma=2.5$.}\label{fig.exp5}
\end{figure*}

\begin{figure}[!t]
	\centering
	\subfigure[\textit{Adult}.]{
	\includegraphics[width=0.48\linewidth]{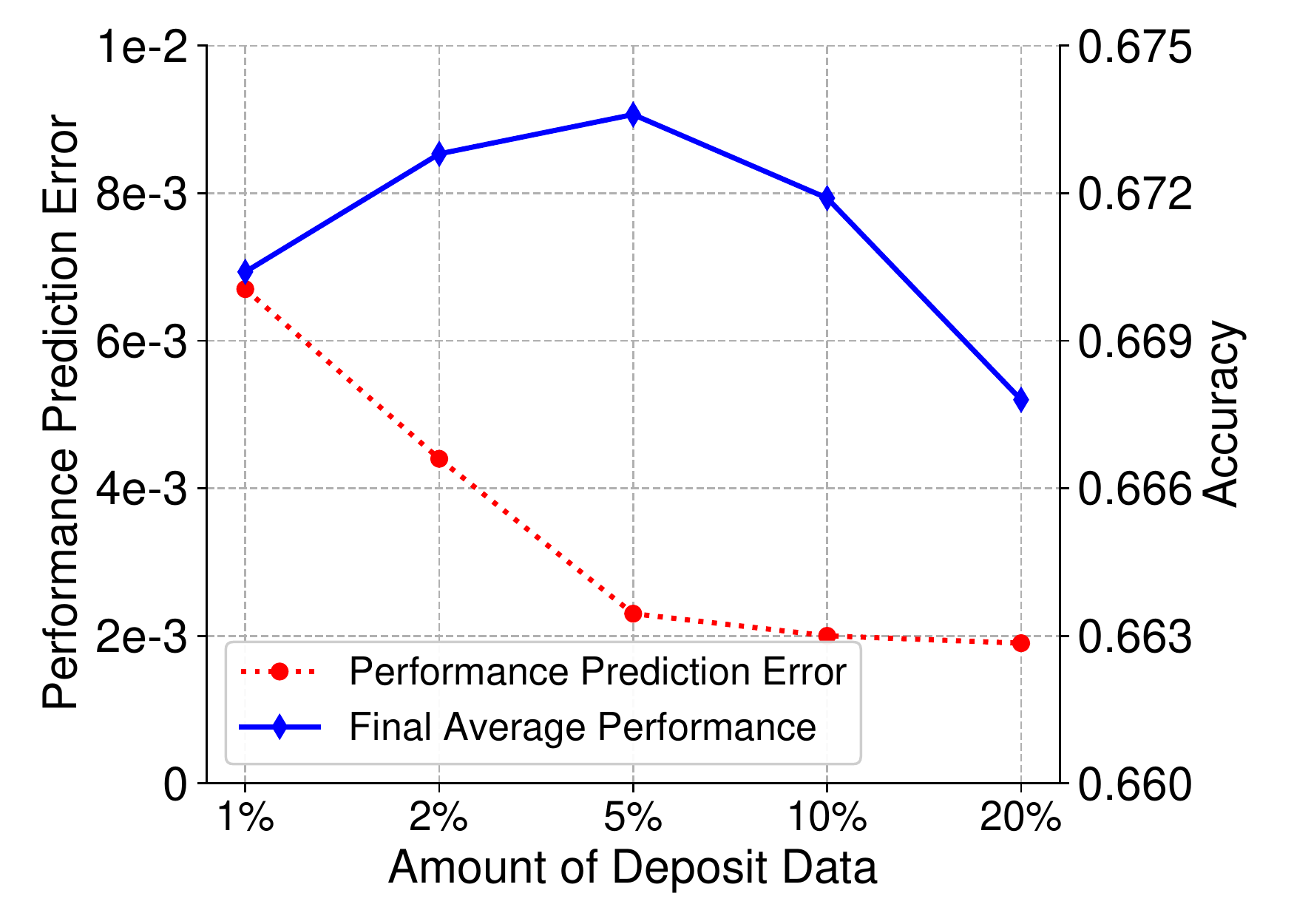}
	}\hspace{-0.05in}
		\subfigure[\textit{CTR}.]{
	\includegraphics[width=0.48\linewidth]{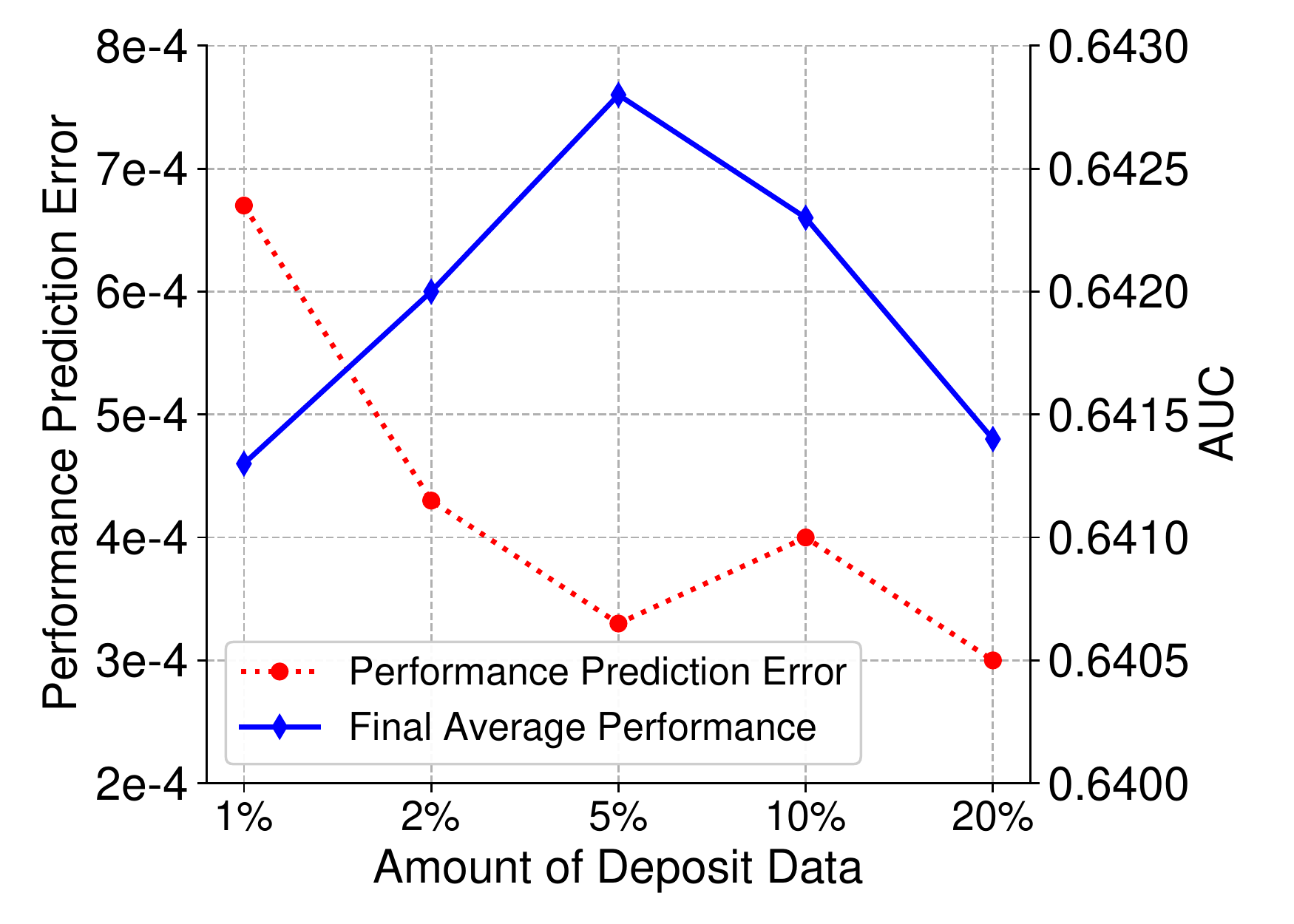}
	}
\caption{Impact of the amount of deposit data on the performance prediction error and the final average performance.}\label{fig.exp6}
\end{figure}

\subsection{Results of Platform Collaboration Policy}

Based on the performance estimation results, we solve the game among different platforms.
To respond to \textbf{RQ3}, we show the final collaboration policies on the two datasets in Fig.~\ref{fig.exp4}.
We have several interesting observations from the policies.
First, we find that all platforms exhaust their privacy budget.
This shows that platforms would like to offer more data to obtain more model performance gains if the privacy budget permits.
In addition, we observe that all platforms tend to collaborate more with the parties that keep more valuable data to their target tasks.
For example, on the \textit{Adult} dataset, the first platform  chooses to offer most of its data to the third platform, since the third platform's data can effectively improve its model performance.
Meanwhile, different from the greedy policy, the third platform still provides a part of its data for the first platform in return, which shows that the platforms have a valid collaboration.
Similar phenomena also exist on the \textit{CTR} dataset.
These results imply that in our method, platforms do not fall into the prisoner's dilemma where all platforms do not provide any data or some platforms are left out from the collaboration.
Our proposed game-based platform negotiation method can facilitate the effective collaborations among different platforms to exploit multi-platform data under privacy restrictions.

\subsection{Hyperparameter Analysis}

To answer \textbf{RQ4}, we illustrate the model accuracy and policy of each platform under different values of $\gamma$, as shown in Fig.~\ref{fig.exp5}.
We find that platforms' policies are more uniform when $\gamma$ is smaller, while are more aggressive when $\gamma$ is larger.
Note that the policy does not change when $\gamma$ is larger than 3 on \textit{Adult} and 3.5 on \textit{CTR}, where both the first and second platforms provide all  their data for the third platform.
To achieve the best collaboration effectiveness among different platforms, a moderate value of $\gamma$ is needed. 
On both datasets, the best average performance is achieved when $\gamma=2.5$, which is consistent with our analysis in the methodology section.

\subsection{Influence of Deposit Data Amount}

To address \textbf{RQ5}, we compare the average platform performance and the average performance prediction errors\footnote{We report the errors under the uniform policy for consistency.} when using different percentages of deposit data, as shown in Fig.~\ref{fig.exp6}.
We find that the performance prediction errors are usually large when the deposit datasets are too small.
This is intuitive because accurate performance estimation usually needs a reasonable amount of data for model learning.
Thus, the average performance is suboptimal if the deposit data is too sparse because the high prediction errors may impair the accuracy of policy optimization.
However, the average performance also declines when the amount of deposit data is too large.
This is because the privacy budget is depleted when a large amount of data is used as the deposit.
Thus, using 5\% of the total data as the deposit is suitable for our method.

\section{Conclusion}\label{sec:Conclusion}

In this paper, we study the platform collaboration problem in vertical federated learning under privacy constraints.
We propose to use reciprocal collaborations to model the incentive mechanisms in VFL, and we further convert it into a  multi-party game problem when considering constraints on platforms' privacy budgets.
To solve this game, we propose a \textit{FedGame} approach that can generate effective collaboration policies.
To evaluate platforms' game rewards under different policies, we propose a performance estimation method that encourages each platform to offer a small amount of deposit data to participate in the federated learning of other platforms.
Each platform independently trains and evaluates multiple models by involving different proportions of deposit data to obtain a regression function between its model performance and the amount of different inter-platform data.
Based on the estimated model performance, each platform participates in a negotiation process to bargain and optimize its collaboration policy.
The final learning is conducted after the policies converge.
Extensive experiments on the two datasets in different scenarios show that our approach can effectively facilitate the collaborations among platforms to better exploit multi-platform data under privacy restrictions.

However, our approach also has the following limitations.
First, in our method, different platforms need to have certain trust, e.g., they do not collude with nor cheat each other.
Second, in our method, we assume that all platforms have the consensus to use the same game solving strategy, which may require additional negotiation before starting the game.
Third, our performance estimation method requires each platform to conduct multiple experiments, which may lead to additional computational and communication costs.
In our future work, we plan to generalize our approach to scenarios with more relaxed assumptions about trust and game strategies.
In addition, we will study more efficient methods to evaluate the value of inter-platform data.

\bibliographystyle{ACM-Reference-Format}
\bibliography{main}


\begin{thebibliography}{46}


\ifx \showCODEN    \undefined \def \showCODEN     #1{\unskip}     \fi
\ifx \showDOI      \undefined \def \showDOI       #1{#1}\fi
\ifx \showISBNx    \undefined \def \showISBNx     #1{\unskip}     \fi
\ifx \showISBNxiii \undefined \def \showISBNxiii  #1{\unskip}     \fi
\ifx \showISSN     \undefined \def \showISSN      #1{\unskip}     \fi
\ifx \showLCCN     \undefined \def \showLCCN      #1{\unskip}     \fi
\ifx \shownote     \undefined \def \shownote      #1{#1}          \fi
\ifx \showarticletitle \undefined \def \showarticletitle #1{#1}   \fi
\ifx \showURL      \undefined \def \showURL       {\relax}        \fi
\providecommand\bibfield[2]{#2}
\providecommand\bibinfo[2]{#2}
\providecommand\natexlab[1]{#1}
\providecommand\showeprint[2][]{arXiv:#2}

\bibitem[Chen et~al\mbox{.}(2020)]%
        {chen2020vafl}
\bibfield{author}{\bibinfo{person}{Tianyi Chen}, \bibinfo{person}{Xiao Jin},
  \bibinfo{person}{Yuejiao Sun}, {and} \bibinfo{person}{Wotao Yin}.}
  \bibinfo{year}{2020}\natexlab{}.
\newblock \showarticletitle{Vafl: a method of vertical asynchronous federated
  learning}.
\newblock \bibinfo{journal}{\emph{arXiv preprint arXiv:2007.06081}}
  (\bibinfo{year}{2020}).
\newblock


\bibitem[Chen and Ali(2002)]%
        {chen2002output}
\bibfield{author}{\bibinfo{person}{Yao Chen} {and} \bibinfo{person}{Agha~Iqbal
  Ali}.} \bibinfo{year}{2002}\natexlab{}.
\newblock \showarticletitle{Output--input ratio analysis and DEA frontier}.
\newblock \bibinfo{journal}{\emph{European Journal of Operational Research}}
  \bibinfo{volume}{142}, \bibinfo{number}{3} (\bibinfo{year}{2002}),
  \bibinfo{pages}{476--479}.
\newblock


\bibitem[Cheng et~al\mbox{.}(2020)]%
        {cheng2020federated}
\bibfield{author}{\bibinfo{person}{Yong Cheng}, \bibinfo{person}{Yang Liu},
  \bibinfo{person}{Tianjian Chen}, {and} \bibinfo{person}{Qiang Yang}.}
  \bibinfo{year}{2020}\natexlab{}.
\newblock \showarticletitle{Federated learning for privacy-preserving AI}.
\newblock \bibinfo{journal}{\emph{Commun. ACM}} \bibinfo{volume}{63},
  \bibinfo{number}{12} (\bibinfo{year}{2020}), \bibinfo{pages}{33--36}.
\newblock


\bibitem[Fan et~al\mbox{.}(2022)]%
        {fan2022fair}
\bibfield{author}{\bibinfo{person}{Zhenan Fan}, \bibinfo{person}{Huang Fang},
  \bibinfo{person}{Zirui Zhou}, \bibinfo{person}{Jian Pei},
  \bibinfo{person}{Michael~P Friedlander}, {and} \bibinfo{person}{Yong Zhang}.}
  \bibinfo{year}{2022}\natexlab{}.
\newblock \showarticletitle{Fair and efficient contribution valuation for
  vertical federated learning}.
\newblock \bibinfo{journal}{\emph{arXiv preprint arXiv:2201.02658}}
  (\bibinfo{year}{2022}).
\newblock


\bibitem[Fang et~al\mbox{.}(2021)]%
        {fang2021large}
\bibfield{author}{\bibinfo{person}{Wenjing Fang}, \bibinfo{person}{Derun Zhao},
  \bibinfo{person}{Jin Tan}, \bibinfo{person}{Chaochao Chen},
  \bibinfo{person}{Chaofan Yu}, \bibinfo{person}{Li Wang}, \bibinfo{person}{Lei
  Wang}, \bibinfo{person}{Jun Zhou}, {and} \bibinfo{person}{Benyu Zhang}.}
  \bibinfo{year}{2021}\natexlab{}.
\newblock \showarticletitle{Large-scale Secure XGB for Vertical Federated
  Learning}. In \bibinfo{booktitle}{\emph{CIKM}}. \bibinfo{pages}{443--452}.
\newblock


\bibitem[Feng et~al\mbox{.}(2019)]%
        {feng2019joint}
\bibfield{author}{\bibinfo{person}{Shaohan Feng}, \bibinfo{person}{Dusit
  Niyato}, \bibinfo{person}{Ping Wang}, \bibinfo{person}{Dong~In Kim}, {and}
  \bibinfo{person}{Ying-Chang Liang}.} \bibinfo{year}{2019}\natexlab{}.
\newblock \showarticletitle{Joint service pricing and cooperative relay
  communication for federated learning}. In \bibinfo{booktitle}{\emph{2019
  International Conference on Internet of Things (iThings) and IEEE Green
  Computing and Communications (GreenCom) and IEEE Cyber, Physical and Social
  Computing (CPSCom) and IEEE Smart Data (SmartData)}}. IEEE,
  \bibinfo{pages}{815--820}.
\newblock


\bibitem[Feng and Yu(2020)]%
        {feng2020multi}
\bibfield{author}{\bibinfo{person}{Siwei Feng} {and} \bibinfo{person}{Han Yu}.}
  \bibinfo{year}{2020}\natexlab{}.
\newblock \showarticletitle{Multi-participant multi-class vertical federated
  learning}.
\newblock \bibinfo{journal}{\emph{arXiv preprint arXiv:2001.11154}}
  (\bibinfo{year}{2020}).
\newblock


\bibitem[Han et~al\mbox{.}(2021)]%
        {han2021data}
\bibfield{author}{\bibinfo{person}{Xiao Han}, \bibinfo{person}{Leye Wang},
  {and} \bibinfo{person}{Junjie Wu}.} \bibinfo{year}{2021}\natexlab{}.
\newblock \showarticletitle{Data Valuation for Vertical Federated Learning: An
  Information-Theoretic Approach}.
\newblock \bibinfo{journal}{\emph{arXiv preprint arXiv:2112.08364}}
  (\bibinfo{year}{2021}).
\newblock


\bibitem[Hard et~al\mbox{.}(2018)]%
        {hard2018federated}
\bibfield{author}{\bibinfo{person}{Andrew Hard}, \bibinfo{person}{Kanishka
  Rao}, \bibinfo{person}{Rajiv Mathews}, \bibinfo{person}{Swaroop Ramaswamy},
  \bibinfo{person}{Fran{\c{c}}oise Beaufays}, \bibinfo{person}{Sean
  Augenstein}, \bibinfo{person}{Hubert Eichner}, \bibinfo{person}{Chlo{\'e}
  Kiddon}, {and} \bibinfo{person}{Daniel Ramage}.}
  \bibinfo{year}{2018}\natexlab{}.
\newblock \showarticletitle{Federated learning for mobile keyboard prediction}.
\newblock \bibinfo{journal}{\emph{arXiv preprint arXiv:1811.03604}}
  (\bibinfo{year}{2018}).
\newblock


\bibitem[Hardy et~al\mbox{.}(2017)]%
        {hardy2017private}
\bibfield{author}{\bibinfo{person}{Stephen Hardy}, \bibinfo{person}{Wilko
  Henecka}, \bibinfo{person}{Hamish Ivey-Law}, \bibinfo{person}{Richard Nock},
  \bibinfo{person}{Giorgio Patrini}, \bibinfo{person}{Guillaume Smith}, {and}
  \bibinfo{person}{Brian Thorne}.} \bibinfo{year}{2017}\natexlab{}.
\newblock \showarticletitle{Private federated learning on vertically
  partitioned data via entity resolution and additively homomorphic
  encryption}.
\newblock \bibinfo{journal}{\emph{arXiv preprint arXiv:1711.10677}}
  (\bibinfo{year}{2017}).
\newblock


\bibitem[Jin et~al\mbox{.}(2021)]%
        {jin2021cafe}
\bibfield{author}{\bibinfo{person}{Xiao Jin}, \bibinfo{person}{Pin-Yu Chen},
  \bibinfo{person}{Chia-Yi Hsu}, \bibinfo{person}{Chia-Mu Yu}, {and}
  \bibinfo{person}{Tianyi Chen}.} \bibinfo{year}{2021}\natexlab{}.
\newblock \showarticletitle{CAFE: Catastrophic Data Leakage in Vertical
  Federated Learning}.
\newblock \bibinfo{journal}{\emph{arXiv preprint arXiv:2110.15122}}
  (\bibinfo{year}{2021}).
\newblock


\bibitem[Kaissis et~al\mbox{.}(2020)]%
        {kaissis2020secure}
\bibfield{author}{\bibinfo{person}{Georgios~A Kaissis},
  \bibinfo{person}{Marcus~R Makowski}, \bibinfo{person}{Daniel R{\"u}ckert},
  {and} \bibinfo{person}{Rickmer~F Braren}.} \bibinfo{year}{2020}\natexlab{}.
\newblock \showarticletitle{Secure, privacy-preserving and federated machine
  learning in medical imaging}.
\newblock \bibinfo{journal}{\emph{Nature Machine Intelligence}}
  \bibinfo{volume}{2}, \bibinfo{number}{6} (\bibinfo{year}{2020}),
  \bibinfo{pages}{305--311}.
\newblock


\bibitem[Kohavi et~al\mbox{.}(1996)]%
        {kohavi1996scaling}
\bibfield{author}{\bibinfo{person}{Ron Kohavi} {et~al\mbox{.}}}
  \bibinfo{year}{1996}\natexlab{}.
\newblock \showarticletitle{Scaling up the accuracy of naive-bayes classifiers:
  A decision-tree hybrid.}. In \bibinfo{booktitle}{\emph{KDD}},
  Vol.~\bibinfo{volume}{96}. \bibinfo{pages}{202--207}.
\newblock


\bibitem[Kone{\v{c}}n{\`y} et~al\mbox{.}(2016)]%
        {konevcny2016federated}
\bibfield{author}{\bibinfo{person}{Jakub Kone{\v{c}}n{\`y}},
  \bibinfo{person}{H~Brendan McMahan}, \bibinfo{person}{Felix~X Yu},
  \bibinfo{person}{Peter Richt{\'a}rik}, \bibinfo{person}{Ananda~Theertha
  Suresh}, {and} \bibinfo{person}{Dave Bacon}.}
  \bibinfo{year}{2016}\natexlab{}.
\newblock \showarticletitle{Federated learning: Strategies for improving
  communication efficiency}.
\newblock \bibinfo{journal}{\emph{arXiv preprint arXiv:1610.05492}}
  (\bibinfo{year}{2016}).
\newblock


\bibitem[Li et~al\mbox{.}(2020)]%
        {li2020federated}
\bibfield{author}{\bibinfo{person}{Tian Li}, \bibinfo{person}{Anit~Kumar Sahu},
  \bibinfo{person}{Ameet Talwalkar}, {and} \bibinfo{person}{Virginia Smith}.}
  \bibinfo{year}{2020}\natexlab{}.
\newblock \showarticletitle{Federated learning: Challenges, methods, and future
  directions}.
\newblock \bibinfo{journal}{\emph{IEEE Signal Processing Magazine}}
  \bibinfo{volume}{37}, \bibinfo{number}{3} (\bibinfo{year}{2020}),
  \bibinfo{pages}{50--60}.
\newblock


\bibitem[Liu et~al\mbox{.}(2020)]%
        {liu2020asymmetrical}
\bibfield{author}{\bibinfo{person}{Yang Liu}, \bibinfo{person}{Xiong Zhang},
  {and} \bibinfo{person}{Libin Wang}.} \bibinfo{year}{2020}\natexlab{}.
\newblock \showarticletitle{Asymmetrical vertical federated learning}.
\newblock \bibinfo{journal}{\emph{arXiv preprint arXiv:2004.07427}}
  (\bibinfo{year}{2020}).
\newblock


\bibitem[Luo et~al\mbox{.}(2021)]%
        {luo2021feature}
\bibfield{author}{\bibinfo{person}{Xinjian Luo}, \bibinfo{person}{Yuncheng Wu},
  \bibinfo{person}{Xiaokui Xiao}, {and} \bibinfo{person}{Beng~Chin Ooi}.}
  \bibinfo{year}{2021}\natexlab{}.
\newblock \showarticletitle{Feature inference attack on model predictions in
  vertical federated learning}. In \bibinfo{booktitle}{\emph{ICDE}}. IEEE,
  \bibinfo{pages}{181--192}.
\newblock


\bibitem[Lyu et~al\mbox{.}(2020)]%
        {lyu2020collaborative}
\bibfield{author}{\bibinfo{person}{Lingjuan Lyu}, \bibinfo{person}{Xinyi Xu},
  \bibinfo{person}{Qian Wang}, {and} \bibinfo{person}{Han Yu}.}
  \bibinfo{year}{2020}\natexlab{}.
\newblock \showarticletitle{Collaborative fairness in federated learning}.
\newblock In \bibinfo{booktitle}{\emph{Federated Learning}}.
  \bibinfo{publisher}{Springer}, \bibinfo{pages}{189--204}.
\newblock


\bibitem[Ma and Tourani(2020)]%
        {ma2020predictive}
\bibfield{author}{\bibinfo{person}{Sisi Ma} {and} \bibinfo{person}{Roshan
  Tourani}.} \bibinfo{year}{2020}\natexlab{}.
\newblock \showarticletitle{Predictive and causal implications of using shapley
  value for model interpretation}. In \bibinfo{booktitle}{\emph{Proceedings of
  the 2020 KDD Workshop on Causal Discovery}}. PMLR, \bibinfo{pages}{23--38}.
\newblock


\bibitem[McMahan et~al\mbox{.}(2017)]%
        {mcmahan2017communication}
\bibfield{author}{\bibinfo{person}{Brendan McMahan}, \bibinfo{person}{Eider
  Moore}, \bibinfo{person}{Daniel Ramage}, \bibinfo{person}{Seth Hampson},
  {and} \bibinfo{person}{Blaise~Aguera y Arcas}.}
  \bibinfo{year}{2017}\natexlab{}.
\newblock \showarticletitle{Communication-Efficient Learning of Deep Networks
  from Decentralized Data}. In \bibinfo{booktitle}{\emph{AISTATS}}.
  \bibinfo{pages}{1273--1282}.
\newblock


\bibitem[Moehle et~al\mbox{.}(2021)]%
        {moehle2021portfolio}
\bibfield{author}{\bibinfo{person}{Nicholas Moehle}, \bibinfo{person}{Stephen
  Boyd}, {and} \bibinfo{person}{Andrew Ang}.} \bibinfo{year}{2021}\natexlab{}.
\newblock \showarticletitle{Portfolio performance attribution via shapley
  value}.
\newblock \bibinfo{journal}{\emph{arXiv preprint arXiv:2102.05799}}
  (\bibinfo{year}{2021}).
\newblock


\bibitem[Ng et~al\mbox{.}(2021)]%
        {ng2021multi}
\bibfield{author}{\bibinfo{person}{Kang~Loon Ng}, \bibinfo{person}{Zichen
  Chen}, \bibinfo{person}{Zelei Liu}, \bibinfo{person}{Han Yu},
  \bibinfo{person}{Yang Liu}, {and} \bibinfo{person}{Qiang Yang}.}
  \bibinfo{year}{2021}\natexlab{}.
\newblock \showarticletitle{A multi-player game for studying federated learning
  incentive schemes}. In \bibinfo{booktitle}{\emph{IJCAI}}.
  \bibinfo{pages}{5279--5281}.
\newblock


\bibitem[Nock et~al\mbox{.}(2018)]%
        {nock2018entity}
\bibfield{author}{\bibinfo{person}{Richard Nock}, \bibinfo{person}{Stephen
  Hardy}, \bibinfo{person}{Wilko Henecka}, \bibinfo{person}{Hamish Ivey-Law},
  \bibinfo{person}{Giorgio Patrini}, \bibinfo{person}{Guillaume Smith}, {and}
  \bibinfo{person}{Brian Thorne}.} \bibinfo{year}{2018}\natexlab{}.
\newblock \showarticletitle{Entity resolution and federated learning get a
  federated resolution}.
\newblock \bibinfo{journal}{\emph{arXiv preprint arXiv:1803.04035}}
  (\bibinfo{year}{2018}).
\newblock


\bibitem[Qi et~al\mbox{.}(2020)]%
        {qi2020privacy}
\bibfield{author}{\bibinfo{person}{Tao Qi}, \bibinfo{person}{Fangzhao Wu},
  \bibinfo{person}{Chuhan Wu}, \bibinfo{person}{Yongfeng Huang}, {and}
  \bibinfo{person}{Xing Xie}.} \bibinfo{year}{2020}\natexlab{}.
\newblock \showarticletitle{Privacy-Preserving News Recommendation Model
  Learning}. In \bibinfo{booktitle}{\emph{EMNLP: Findings}}.
  \bibinfo{pages}{1423--1432}.
\newblock


\bibitem[Rieke et~al\mbox{.}(2020)]%
        {rieke2020future}
\bibfield{author}{\bibinfo{person}{Nicola Rieke}, \bibinfo{person}{Jonny
  Hancox}, \bibinfo{person}{Wenqi Li}, \bibinfo{person}{Fausto Milletari},
  \bibinfo{person}{Holger~R Roth}, \bibinfo{person}{Shadi Albarqouni},
  \bibinfo{person}{Spyridon Bakas}, \bibinfo{person}{Mathieu~N Galtier},
  \bibinfo{person}{Bennett~A Landman}, \bibinfo{person}{Klaus Maier-Hein},
  {et~al\mbox{.}}} \bibinfo{year}{2020}\natexlab{}.
\newblock \showarticletitle{The future of digital health with federated
  learning}.
\newblock \bibinfo{journal}{\emph{NPJ digital medicine}} \bibinfo{volume}{3},
  \bibinfo{number}{1} (\bibinfo{year}{2020}), \bibinfo{pages}{1--7}.
\newblock


\bibitem[Roy~Choudhary et~al\mbox{.}(2014)]%
        {roy2014cross}
\bibfield{author}{\bibinfo{person}{Shauvik Roy~Choudhary},
  \bibinfo{person}{Mukul~R Prasad}, {and} \bibinfo{person}{Alessandro Orso}.}
  \bibinfo{year}{2014}\natexlab{}.
\newblock \showarticletitle{Cross-platform feature matching for web
  applications}. In \bibinfo{booktitle}{\emph{ISSTA}}. \bibinfo{pages}{82--92}.
\newblock


\bibitem[Song et~al\mbox{.}(2019)]%
        {song2019profit}
\bibfield{author}{\bibinfo{person}{Tianshu Song}, \bibinfo{person}{Yongxin
  Tong}, {and} \bibinfo{person}{Shuyue Wei}.} \bibinfo{year}{2019}\natexlab{}.
\newblock \showarticletitle{Profit allocation for federated learning}. In
  \bibinfo{booktitle}{\emph{Big Data}}. IEEE, \bibinfo{pages}{2577--2586}.
\newblock


\bibitem[Voigt and Von~dem Bussche(2017)]%
        {voigt2017eu}
\bibfield{author}{\bibinfo{person}{Paul Voigt} {and} \bibinfo{person}{Axel
  Von~dem Bussche}.} \bibinfo{year}{2017}\natexlab{}.
\newblock \showarticletitle{The eu general data protection regulation (gdpr)}.
\newblock \bibinfo{journal}{\emph{A Practical Guide, 1st Ed., Cham: Springer
  International Publishing}}  \bibinfo{volume}{10} (\bibinfo{year}{2017}),
  \bibinfo{pages}{3152676}.
\newblock


\bibitem[Wang(2019)]%
        {wang2019interpret}
\bibfield{author}{\bibinfo{person}{Guan Wang}.}
  \bibinfo{year}{2019}\natexlab{}.
\newblock \showarticletitle{Interpret federated learning with shapley values}.
\newblock \bibinfo{journal}{\emph{arXiv preprint arXiv:1905.04519}}
  (\bibinfo{year}{2019}).
\newblock


\bibitem[Wang et~al\mbox{.}(2019)]%
        {wang2019measure}
\bibfield{author}{\bibinfo{person}{Guan Wang},
  \bibinfo{person}{Charlie~Xiaoqian Dang}, {and} \bibinfo{person}{Ziye Zhou}.}
  \bibinfo{year}{2019}\natexlab{}.
\newblock \showarticletitle{Measure contribution of participants in federated
  learning}. In \bibinfo{booktitle}{\emph{Big Data}}. IEEE,
  \bibinfo{pages}{2597--2604}.
\newblock


\bibitem[Wang et~al\mbox{.}(2020)]%
        {wang2020principled}
\bibfield{author}{\bibinfo{person}{Tianhao Wang}, \bibinfo{person}{Johannes
  Rausch}, \bibinfo{person}{Ce Zhang}, \bibinfo{person}{Ruoxi Jia}, {and}
  \bibinfo{person}{Dawn Song}.} \bibinfo{year}{2020}\natexlab{}.
\newblock \showarticletitle{A principled approach to data valuation for
  federated learning}.
\newblock In \bibinfo{booktitle}{\emph{Federated Learning}}.
  \bibinfo{publisher}{Springer}, \bibinfo{pages}{153--167}.
\newblock


\bibitem[Wu et~al\mbox{.}(2021)]%
        {wu2021fedgnn}
\bibfield{author}{\bibinfo{person}{Chuhan Wu}, \bibinfo{person}{Fangzhao Wu},
  \bibinfo{person}{Yang Cao}, \bibinfo{person}{Yongfeng Huang}, {and}
  \bibinfo{person}{Xing Xie}.} \bibinfo{year}{2021}\natexlab{}.
\newblock \showarticletitle{Fedgnn: Federated graph neural network for
  privacy-preserving recommendation}.
\newblock \bibinfo{journal}{\emph{arXiv preprint arXiv:2102.04925}}
  (\bibinfo{year}{2021}).
\newblock


\bibitem[Wu et~al\mbox{.}(2022)]%
        {wu2020fedctr}
\bibfield{author}{\bibinfo{person}{Chuhan Wu}, \bibinfo{person}{Fangzhao Wu},
  \bibinfo{person}{Lingjuan Lyu}, \bibinfo{person}{Yongfeng Huang}, {and}
  \bibinfo{person}{Xing Xie}.} \bibinfo{year}{2022}\natexlab{}.
\newblock \showarticletitle{FedCTR: Federated Native Ad CTR Prediction with
  Cross Platform User Behavior Data}.
\newblock \bibinfo{journal}{\emph{ACM Transactions on Intelligent Systems and
  Technology (TIST)}} (\bibinfo{year}{2022}).
\newblock


\bibitem[Wu and Soo(1999)]%
        {WuS99}
\bibfield{author}{\bibinfo{person}{Shih{-}Hung Wu} {and}
  \bibinfo{person}{Von{-}Wun Soo}.} \bibinfo{year}{1999}\natexlab{}.
\newblock \showarticletitle{Game Theoretic Reasoning in Multi-Agent
  Coordination by Negotiation with a Trusted Third Party}. In
  \bibinfo{booktitle}{\emph{AGENTS}}, \bibfield{editor}{\bibinfo{person}{Oren
  Etzioni}, \bibinfo{person}{J{\"{o}}rg~P. M{\"{u}}ller}, {and}
  \bibinfo{person}{Jeffrey~M. Bradshaw}} (Eds.). \bibinfo{publisher}{ACM},
  \bibinfo{pages}{56--61}.
\newblock


\bibitem[Wu et~al\mbox{.}(2020)]%
        {wu2020privacy}
\bibfield{author}{\bibinfo{person}{Yuncheng Wu}, \bibinfo{person}{Shaofeng
  Cai}, \bibinfo{person}{Xiaokui Xiao}, \bibinfo{person}{Gang Chen}, {and}
  \bibinfo{person}{Beng~Chin Ooi}.} \bibinfo{year}{2020}\natexlab{}.
\newblock \showarticletitle{Privacy preserving vertical federated learning for
  tree-based models}.
\newblock \bibinfo{journal}{\emph{VLDB}} \bibinfo{volume}{13},
  \bibinfo{number}{12} (\bibinfo{year}{2020}), \bibinfo{pages}{2090--2103}.
\newblock


\bibitem[Yan et~al\mbox{.}(2021)]%
        {yan2021fedcm}
\bibfield{author}{\bibinfo{person}{Bingjie Yan}, \bibinfo{person}{Boyi Liu},
  \bibinfo{person}{Lujia Wang}, \bibinfo{person}{Yize Zhou},
  \bibinfo{person}{Zhixuan Liang}, \bibinfo{person}{Ming Liu}, {and}
  \bibinfo{person}{Cheng{-}Zhong Xu}.} \bibinfo{year}{2021}\natexlab{}.
\newblock \showarticletitle{FedCM: {A} Real-time Contribution Measurement
  Method for Participants in Federated Learning}. In
  \bibinfo{booktitle}{\emph{IJCNN}}. \bibinfo{pages}{1--8}.
\newblock


\bibitem[Yang et~al\mbox{.}(2019a)]%
        {yang2019quasi}
\bibfield{author}{\bibinfo{person}{Kai Yang}, \bibinfo{person}{Tao Fan},
  \bibinfo{person}{Tianjian Chen}, \bibinfo{person}{Yuanming Shi}, {and}
  \bibinfo{person}{Qiang Yang}.} \bibinfo{year}{2019}\natexlab{a}.
\newblock \showarticletitle{A Quasi-Newton Method Based Vertical Federated
  Learning Framework for Logistic Regression}.
\newblock \bibinfo{journal}{\emph{arXiv preprint arXiv:1912.00513}}
  (\bibinfo{year}{2019}).
\newblock


\bibitem[Yang et~al\mbox{.}(2020)]%
        {yang2020federated}
\bibfield{author}{\bibinfo{person}{Liu Yang}, \bibinfo{person}{Ben Tan},
  \bibinfo{person}{Vincent~W Zheng}, \bibinfo{person}{Kai Chen}, {and}
  \bibinfo{person}{Qiang Yang}.} \bibinfo{year}{2020}\natexlab{}.
\newblock \showarticletitle{Federated recommendation systems}.
\newblock In \bibinfo{booktitle}{\emph{Federated Learning}}.
  \bibinfo{publisher}{Springer}, \bibinfo{pages}{225--239}.
\newblock


\bibitem[Yang et~al\mbox{.}(2019b)]%
        {yang2019federated}
\bibfield{author}{\bibinfo{person}{Qiang Yang}, \bibinfo{person}{Yang Liu},
  \bibinfo{person}{Tianjian Chen}, {and} \bibinfo{person}{Yongxin Tong}.}
  \bibinfo{year}{2019}\natexlab{b}.
\newblock \showarticletitle{Federated machine learning: Concept and
  applications}.
\newblock \bibinfo{journal}{\emph{TIST}} \bibinfo{volume}{10},
  \bibinfo{number}{2} (\bibinfo{year}{2019}), \bibinfo{pages}{1--19}.
\newblock


\bibitem[Yang et~al\mbox{.}(2019c)]%
        {yang2019parallel}
\bibfield{author}{\bibinfo{person}{Shengwen Yang}, \bibinfo{person}{Bing Ren},
  \bibinfo{person}{Xuhui Zhou}, {and} \bibinfo{person}{Liping Liu}.}
  \bibinfo{year}{2019}\natexlab{c}.
\newblock \showarticletitle{Parallel distributed logistic regression for
  vertical federated learning without third-party coordinator}.
\newblock \bibinfo{journal}{\emph{arXiv preprint arXiv:1911.09824}}
  (\bibinfo{year}{2019}).
\newblock


\bibitem[Yu et~al\mbox{.}(2020)]%
        {yu2020fairness}
\bibfield{author}{\bibinfo{person}{Han Yu}, \bibinfo{person}{Zelei Liu},
  \bibinfo{person}{Yang Liu}, \bibinfo{person}{Tianjian Chen},
  \bibinfo{person}{Mingshu Cong}, \bibinfo{person}{Xi Weng},
  \bibinfo{person}{Dusit Niyato}, {and} \bibinfo{person}{Qiang Yang}.}
  \bibinfo{year}{2020}\natexlab{}.
\newblock \showarticletitle{A fairness-aware incentive scheme for federated
  learning}. In \bibinfo{booktitle}{\emph{AIES}}. \bibinfo{pages}{393--399}.
\newblock


\bibitem[Zeng et~al\mbox{.}(2021)]%
        {zeng2021comprehensive}
\bibfield{author}{\bibinfo{person}{Rongfei Zeng}, \bibinfo{person}{Chao Zeng},
  \bibinfo{person}{Xingwei Wang}, \bibinfo{person}{Bo Li}, {and}
  \bibinfo{person}{Xiaowen Chu}.} \bibinfo{year}{2021}\natexlab{}.
\newblock \showarticletitle{A Comprehensive Survey of Incentive Mechanism for
  Federated Learning}.
\newblock \bibinfo{journal}{\emph{arXiv preprint arXiv:2106.15406}}
  (\bibinfo{year}{2021}).
\newblock


\bibitem[Zhan et~al\mbox{.}(2020)]%
        {zhan2020learning}
\bibfield{author}{\bibinfo{person}{Yufeng Zhan}, \bibinfo{person}{Peng Li},
  \bibinfo{person}{Zhihao Qu}, \bibinfo{person}{Deze Zeng}, {and}
  \bibinfo{person}{Song Guo}.} \bibinfo{year}{2020}\natexlab{}.
\newblock \showarticletitle{A learning-based incentive mechanism for federated
  learning}.
\newblock \bibinfo{journal}{\emph{IEEE IoTJ}} \bibinfo{volume}{7},
  \bibinfo{number}{7} (\bibinfo{year}{2020}), \bibinfo{pages}{6360--6368}.
\newblock


\bibitem[Zhang et~al\mbox{.}(2021b)]%
        {zhang2021survey}
\bibfield{author}{\bibinfo{person}{Chen Zhang}, \bibinfo{person}{Yu Xie},
  \bibinfo{person}{Hang Bai}, \bibinfo{person}{Bin Yu},
  \bibinfo{person}{Weihong Li}, {and} \bibinfo{person}{Yuan Gao}.}
  \bibinfo{year}{2021}\natexlab{b}.
\newblock \showarticletitle{A survey on federated learning}.
\newblock \bibinfo{journal}{\emph{Knowledge-Based Systems}}
  \bibinfo{volume}{216} (\bibinfo{year}{2021}), \bibinfo{pages}{106775}.
\newblock


\bibitem[Zhang et~al\mbox{.}(2021a)]%
        {zhang2021secure}
\bibfield{author}{\bibinfo{person}{Qingsong Zhang}, \bibinfo{person}{Bin Gu},
  \bibinfo{person}{Cheng Deng}, {and} \bibinfo{person}{Heng Huang}.}
  \bibinfo{year}{2021}\natexlab{a}.
\newblock \showarticletitle{Secure Bilevel Asynchronous Vertical Federated
  Learning with Backward Updating}. In \bibinfo{booktitle}{\emph{AAAI}},
  Vol.~\bibinfo{volume}{35}. \bibinfo{pages}{10896--10904}.
\newblock


\bibitem[Zhou et~al\mbox{.}(2021)]%
        {zhou2021towards}
\bibfield{author}{\bibinfo{person}{Zirui Zhou}, \bibinfo{person}{Lingyang Chu},
  \bibinfo{person}{Changxin Liu}, \bibinfo{person}{Lanjun Wang},
  \bibinfo{person}{Jian Pei}, {and} \bibinfo{person}{Yong Zhang}.}
  \bibinfo{year}{2021}\natexlab{}.
\newblock \showarticletitle{Towards Fair Federated Learning}. In
  \bibinfo{booktitle}{\emph{KDD}}. \bibinfo{pages}{4100--4101}.
\newblock


\end{thebibliography}

\newpage
\appendix
\section{Supplementary Materials}

\subsection{Feature Field Partition}

We list the feature fields on different simulated platforms as well as the features used as the prediction targets in Table~\ref{table.s1}.
The feature orders are consistent with the original orders in the dataset.
Note that the income feature is a binary variable that indicates whether the income of a person is higher than 50K.

\begin{table}[h]
\caption{The platform partition of feature fields.}\label{table.s1}
\begin{tabular}{cllc}
\Xhline{1.0pt}
\textbf{Platform}  & \multicolumn{1}{c}{\textbf{Feature}} & \multicolumn{1}{c}{\textbf{Type}} & \textbf{Notes} \\ \hline
\multirow{5}{*}{1} & age                                  & continuous                        &                \\
                   & workclass                            & categorical                       &                \\
                   & final-weight                         & continuous                        &                \\
                   & education                            & categorical                       & Label          \\
                   & education-num                        & continuous                        & Removed        \\ \hline
\multirow{5}{*}{2} & marital-status                       & categorical                       &                \\
                   & occupation                           & categorical                       &                \\
                   & relationship                         & categorical                       &                \\
                   & race                                 & categorical                       &                \\
                   & gender                               & categorical                       & Label          \\ \hline
\multirow{5}{*}{3} & capital-gain                         & continuous                        &                \\
                   & capital-loss                         & continuous                        &                \\
                   & hours-per-week                       & continuous                        &                \\
                   & native-country                       & categorical                       &                \\
                   & income                               &    categorical                               & Label          \\ \Xhline{1.0pt}
\end{tabular}
\end{table}

\subsection{Experimental Environment}

Our experiments are conducted on a Linux machine with Ubuntu 16.04 operating system.
The codes are written in Python 3.8.
The deep learning algorithms are implemented by the Keras library 2.2.4 with Tensorflow 1.15 backend.
On the \textit{Adult} dataset, we only use CPU to run experiments.
On the \textit{CTR} dataset, we run experiment on a Nvidia Tesla V100 GPU with 32GB memory.
Note that the negotiation process takes little computing resource and the game can be solved in seconds with common PCs.

\subsection{Preprocessing}

On the \textit{Adult} dataset, we use one-hot encoding to map categorical features, and we normalize other numerical features.
To fill missing values, we use the mean values for numerical features and the majority class for categorical features.
On the \textit{CTR} dataset, we use the NLTK tool to preprocess the texts.
We use the word\_tokenize function to convert the input texts into token sequences.
We filter the low frequency tokens in search queries and browsed webpages using a frequency threshold of 10.
The token embeddings of out-of-vocabulary words are filled with random vectors that have the same mean and co-variation values as other tokens.

\end{document}